\documentclass[runningheads]{llncs}
\usepackage{graphicx}
\usepackage{comment}
\usepackage{amsmath,amssymb} %
\usepackage[table,dvipsnames]{xcolor}
\usepackage{symbols}
\usepackage{xspace}
\usepackage{arydshln}
\usepackage{tabularx}
\usepackage{gensymb}
\usepackage{pifont}
\usepackage[normalem]{ulem}
\usepackage{soul}
\colorlet{greenhl}{green!15}
\usepackage{multirow}
\usepackage[shortlabels]{enumitem}
\usepackage[colorlinks]{hyperref}       
\usepackage{subcaption}

\newcommand{\oldtask}{\mbox{\sc{FurnLift}}\xspace}
\newcommand{\task}{\mbox{\sc{FurnMove}}\xspace}
\newcommand{\gridtask}{\mbox{Gridworld-\sc{FurnMove}}\xspace}
\newcommand{\method}{\mbox{\sc{SYNC}}\xspace}
\newcommand{\itmethod}{\emph{SYNC}\xspace}
\newcommand{\loss}{\mbox{\sc{CORDIAL}}\xspace}
\newcommand{\thor}{AI2-THOR\xspace}

\sethlcolor{greenhl}

\newcommand{\myspl}{\textit{MD-SPL}\xspace}
\newcommand{\eplen}{\textit{Ep~len}\xspace}
\newcommand{\srate}{\textit{Success}\xspace}
\newcommand{\fdist}{\textit{Final dist}\xspace}

\newcommand{\lrankdist}{\textit{TVD}\xspace}
\newcommand{\invalidprob}{\textit{Invalid~prob}\xspace}

\newcommand{\navactions}{\mathcal{A}^{\text{NAV}}}
\newcommand{\mwoactions}{\mathcal{A}^{\textsc{MWO}}}
\newcommand{\moactions}{\mathcal{A}^{\textsc{MO}}}
\newcommand{\roactions}{\mathcal{A}^{\textsc{RO}}}
\newif\ifisarxiv
\isarxivfalse
\newcommand{\appendixorsupp}{\ifisarxiv{appendix}\else{supplement}\fi\xspace}

\begin{document}
\pagestyle{headings}
\mainmatter
\def\ECCVSubNumber{3663}  %

\title{A Cordial Sync: Going Beyond Marginal Policies for Multi-Agent Embodied Tasks} %

\titlerunning{\textsc{A Cordial Sync}}
\author{
Unnat Jain\inst{1}\thanks{denotes equal contribution by UJ and LW}
\and
Luca Weihs\inst{2}$^\star$
\and
Eric Kolve\inst{2}
\and
Ali Farhadi\inst{3}
\and\\
Svetlana Lazebnik\inst{1}
\and
Aniruddha Kembhavi\inst{2,3}
\and
Alexander Schwing\inst{1}
}
\authorrunning{U. Jain \& L. Weihs et al.}
\institute{
University of Illinois at Urbana-Champaign\\ \and Allen Institute for AI \and University of Washington
}

\maketitle
\begin{abstract}

Autonomous agents must learn to collaborate. It is not scalable to develop a new centralized agent every time a task's difficulty outpaces a single agent's abilities. While multi-agent collaboration research has flourished in gridworld-like environments, relatively little work has considered visually rich domains. Addressing this, we introduce the novel task \task in which agents work together to move a piece of furniture through a living room to a goal. Unlike existing tasks, \task requires agents to coordinate at every timestep. We identify two challenges when training agents to complete \task: existing decentralized action sampling procedures do not permit expressive joint action policies and, in tasks requiring close coordination, the number of failed actions dominates successful actions. To confront these challenges we introduce \method-policies (synchronize your actions coherently) and \loss (coordination loss). Using \method-policies and \loss, our agents achieve a 58\% completion rate on \task, an impressive absolute gain of 25 percentage points over competitive decentralized baselines. Our dataset, code, and pretrained models are available at \url{https://unnat.github.io/cordial-sync}.
\keywords{Embodied agents, multi-agent reinforcement learning, collaboration, emergent communication, AI2-THOR}
\vspace{-6mm}
\end{abstract}

\begin{figure}[t]
    \centering
    \includegraphics[width=1.0\textwidth]{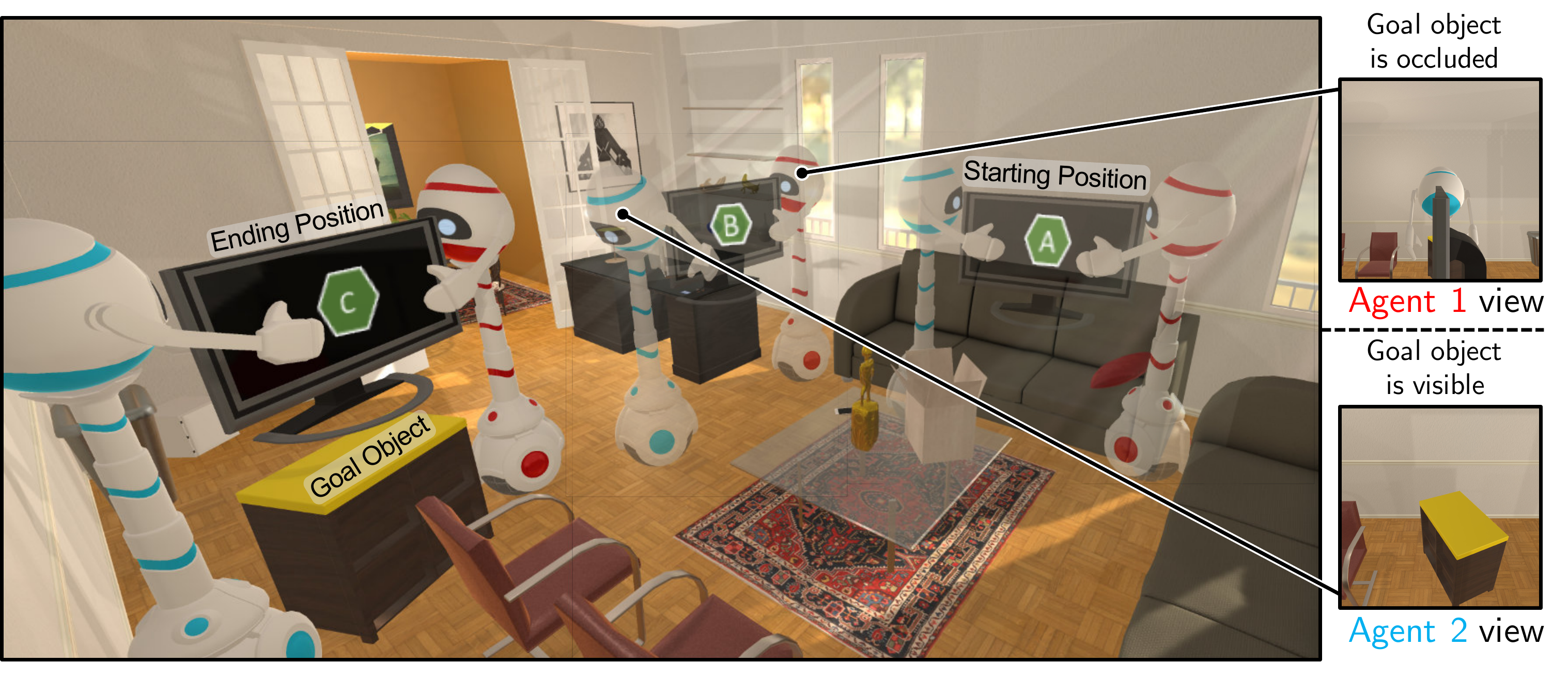}
    \caption{Two agents communicate and synchronize their actions to move a heavy object through a complex indoor environment towards a goal. (a) Agents are initialized holding the object in a randomly chosen location. (b) Note the agent's egocentric views. Successful navigation requires agents to  communicate their intent to reposition themselves, and the object, while contending with collisions, mutual occlusion, and partial information. (c) Agents successfully moved the object above the goal}
    \label{fig:teaser}
\end{figure}

\section{Introduction}
\label{sec:introduction}
Collaboration is the defining principle of our society. %
Humans have refined strategies to efficiently collaborate, developing verbal, deictic, and kinesthetic means. In contrast, progress towards enabling artificial embodied agents to learn collaborative strategies is still in its infancy. Prior work mostly studies collaborative agents in grid-world like environments. Visual, multi-agent, collaborative tasks have not been studied until very recently~\cite{das2018tarmac,jain2019CVPRTBONE}. While existing tasks are well designed to study some aspects of collaboration, they often don't require  agents to closely collaborate \emph{throughout} the task. Instead such tasks either require initial coordination (distributing tasks) followed by almost independent execution, or collaboration at a task's end (\eg, verifying completion). Few tasks require  frequent coordination, and we are aware of none  within a visual setting.

To study our algorithmic ability to address tasks which require close and frequent collaboration, 
we introduce the furniture moving (\task) task (see Fig.~\ref{fig:teaser}), set in the  %
\thor environment. Given only their egocentric visual observations, agents jointly hold a lifted piece of furniture in a living room scene and must collaborate to move it to a visually distinct goal location. %
As a piece of furniture cannot be moved without both agents agreeing on the direction, agents must explicitly \emph{coordinate at every timestep}. Beyond coordinating actions, high performance in our task requires agents to visually anticipate possible collisions, handle occlusion due to obstacles and other agents, and estimate free space. %
Akin to the challenges faced by a group of roommates relocating a widescreen television, this task necessitates extensive and ongoing coordination  amongst all agents at every time step. 

In prior work, collaboration between multiple agents has been enabled primarily by (i) sharing observations or (ii) learning low-bandwidth communication. (i) is often implemented using a \emph{centralized} agent, \ie, a single agent with access to all observations from all agents~\cite{boutilier1999sequential,peng2017multiagent,usunier2016episodic}. While effective it is also unrealistic: the real world poses restrictions on communication bandwidth, latency, and modality. %
We are interested in the more realistic \emph{decentralized} setting enabled via option (ii). %
This is often 
implemented by one or more rounds of message passing between agents before they choose their actions~\cite{FoersterNIPS2016,LoweNIPS2017,jain2019CVPRTBONE}. %
Training decentralized agents when faced with \task's requirement of coordination at each timestep leads to two technical challenges. Challenge 1: as each agent independently samples an action from its policy at every timestep, the joint probability tensor of all agents' actions at any given time is  rank-one. This severely limits which multi-agent policies are representable. %
Challenge 2: the number of possible mis-steps or failed actions increases dramatically when requiring that agents closely coordinate with each other, complicating training. %

Addressing challenge 1, we introduce \method{} (\textbf{S}ynchronize \textbf{Y}our actio\textbf{N}s \textbf{C}oherently) policies 
which permit expressive (\ie, beyond rank-one) joint policies for decentralized agents while using interpretable communication. %
To ameliorate challenge 2 we introduce the \textbf{C}o\textbf{ordi}n\textbf{a}tion \textbf{L}oss (\loss) that replaces the standard entropy  loss in actor-critic algorithms and guides agents away from  actions that are mutually incompatible. %
A 2-agent system using \method and \loss obtains a 58\% success rate on test scenes in \task, an impressive absolute gain of 25 %
percentage points %
over the baseline from \cite{jain2019CVPRTBONE} (76\% relative gain). In a 3-agent setting, this difference is even more extreme.

In summary, our contributions are: (i) \task, a new multi-agent embodied task that demands  ongoing coordination, (ii) \method, a collaborative mechanism that permits expressive joint action policies for decentralized agents, (iii) \loss, a training loss for multi-agent setups which, when combined with \method, leads to large gains, and (iv) improvements to the open-source AI2-THOR environment including a $16\times$ faster gridworld equivalent enabling fast prototyping. 

\section{Related work}
\label{sec:related_work}
We start by reviewing single agent embodied AI tasks followed by non-visual Multi-Agent RL (MARL) and end with visual MARL.

\noindent\textbf{Single-agent embodied systems:} Single-agent embodied systems have been considered extensively in the literature. For instance,  
literature on visual navigation, \ie, locating an object of interest given only visual input, %
spans geometric and learning based methods. 
Geometric approaches have been proposed separately for mapping and planning phases of navigation. Methods entailing structure-from-motion and SLAM~\cite{tomasi1992shape,frahm2016structurefrommotion,thorpe2000structure,cadena2016past,smith1986on,smith1986estimating} were used to build maps. Planning algorithms on existing maps~\cite{CannyMIT1988,KavrakiRA1996,Lavalle2000} and combined mapping \& planning~\cite{elfes1989using,kuipers1991byun,konolige2010viewbased,fraundorfer2012visionbased,aydemir2013active} are other related research directions. 

While these works propose geometric approaches, the task of navigation can be cast as a reinforcement learning (RL) problem, mapping pixels to policies in an end-to-end manner.  RL approaches~\cite{oh2016control,abel2016exploratory,daftry2016learning,giusti2016he,kahn2017plato,toussaint2003learning,mirowski2017learning,tamar2016value} have been proposed to address navigation in synthetic layouts like mazes, arcade games and other visual  environments~\cite{wymann2013torcs,bellemare2013the,kempka2016vizdoom,lerer2016learning,johnson2016the,SukhbaatarARXIV2015}. %
Navigation within photo-realistic environments%
~\cite{BrodeurARXIV2017,SavvaARXIV2017Minos,Chang3DV2017Matterport,ai2thor,xia2018gibson,stanford2d3d,GuptaCVPR2018,xia2019interactive,habitat19iccv} led development of \textit{embodied} AI agents. The early work~\cite{ZhuARXIV2016} addressed object navigation (find an object given an image) in \thor. Soon after,~\cite{GuptaCVPR2018} showed how imitation learning permits agents to learn to build a map from which they navigate. Methods also investigate the utility of topological and latent memory maps~\cite{GuptaCVPR2018,savinov2018semiparametric,henriques2018mapnet,wu2019bayesian}, graph-based learning~\cite{wu2019bayesian,yang2018visual}, meta-learning~\cite{wortsman2019learning}, unimodal baselines~\cite{thomason2019shifting}, 3D point clouds~\cite{Wijmans2019EQAPhoto}, and effective exploration~\cite{wang2019reinforced,savinov2018semiparametric,Chaplot2020Explore,ramakrishnan2020exploration} to improve embodied navigational agents. Embodied navigation also aids AI agents to develop behavior such as instruction following~\cite{HillARXIV2017,anderson2018vision,Suhr2019CerealBar,wang2019reinforced,anderson2019NeuripsChasing}, city navigation~\cite{chen2019touchdown,mirowski2018learningcity,mirowski2019streetlearn,de2018talkthewalk}, question answering~\cite{DasCVPR2018,DasECCV2018,GordonCVPR2018,Wijmans2019EQAPhoto,das2020probing}, and active visual recognition~\cite{yang2018visualsemantic,yang2019embodied}. Recently, with visual and acoustic rendering, agents have been trained for audio-visual embodied navigation~\cite{chen2019audio,gao2020visualechoes}.

In contrast to the above single-agent embodied tasks and approaches, we focus on collaboration between multiple embodied agents. Porting the above single-agent architectural novelties (or a combination of them) to multi-agent systems such as ours is an interesting direction for future work.

\noindent\textbf{Non-visual MARL:} 
Multi-agent reinforcement learning (MARL) is challenging due to non-stationarity when learning. Multiple methods have been proposed to address resulting issues~\cite{TanICML1993,TesauroNIPS2004,TampuuPLOS2017,FoersterARXIV2017}. For instance, permutation invariant critics have been developed recently~\cite{LiuCORL2019}. In addition, for MARL, cooperation and  competition between agents has been studied~\cite{LauerICML2000,Panait2005,MatignonIROS2007,Busoniu2008,OmidshafieiARXIV2017,GuptaAAMAS2017,LoweNIPS2017,FoersterAAAI2018,LiuCORL2019}. 
Similarly, communication and language in the multi-agent setting has been investigated~\cite{GilesICABS2002,KasaiSCIA2008,BratmanCogMod2010,MeloMAS2011,LazaridouARXIV2016,FoersterNIPS2016,SukhbaatarNIPS2016,MordatchAAAI2018,Baker2019EmergentTU} in maze-based setups, tabular tasks, or Markov games. These algorithms mostly operate on low-dimensional observations such as kinematic measurements (position, velocity, \etc) and top-down occupancy grids. For a survey of centralized and decentralized MARL methods, kindly refer to~\cite{zhang2019multi}.
Our work differs from the aforementioned MARL works in that we consider complex visual environments. Our contribution of SYNC-Policies is largely orthogonal to RL loss function or method. For a fair comparison to~\cite{jain2019CVPRTBONE}, we used the same RL algorithm (A3C) but it is straightforward to integrate SYNC into other MARL methods~\cite{rashid2018qmix,FoersterAAAI2018,LoweNIPS2017} (for details, see~\secref{sec:extra-training-details} of the supplement).

\noindent\textbf{Visual MARL:} 
Recently, Jain \etal~\cite{jain2019CVPRTBONE} introduced a collaborative task for two embodied visual agents, which we refer as \oldtask. In this task, two agents are randomly initialized in an \thor living room scene, must visually navigate to a TV, and, in a singe coordinated \textsc{PickUp} action, work to lift that TV up. Note that \oldtask doesn't demand that agents coordinate their actions at each timestep. Instead, such coordination only occurs at the last timestep of an episode. Moreover, as  success of an action executed by an agent is independent (with the exception of the \textsc{PickUp} action), a high performance joint policy need not be complex, \ie, it may be near low-rank. More details on this analysis and the complexity of our proposed \task task are provided in \secref{sec:task}.

Similarly, a recent preprint~\cite{chen2019visual} proposes a visual hide-and-seek task, where agents can move independently. Das~\etal~\cite{das2018tarmac} enable agents to learn who to communicate with, on predominantly 2D tasks. In visual environments they study the task where multiple agents parallely navigate to the same object. Jaderberg~\etal~\cite{jaderberg2019human}  recently studied the game of Quake III  and Weihs~\etal~\cite{weihs2019artificial}  develop agents to play an adversarial hiding game in \thor. Collaborative perception for semantic segmentation and recognition classification have also been investigated recently~\cite{Liu_2020_CVPR,liu2020who2com}.

To the best of our knowledge, all previous visual or non-visual MARL in the decentralized setting operate with a single marginal probability distribution per agent, \ie, a rank-one joint distribution.
Moreover, \task is the first multi-agent collaborative task in a visually rich domain requiring close coordination between agents at every timestep.

\section{The furniture moving task (\task)} \label{sec:task}

We describe our new multi-agent task \task, grounded in the real-world experience of moving furniture. We begin by introducing notation.

\noindent\textbf{RL background and notation.}
Consider $N\geq 1$ collaborative agents $A^1$, $\ldots,$ $A^N$. At every timestep $t\in\mathbb{N}=\{0,1,\ldots\}$ the agents, and environment, are
in some state $s_t\in\mathcal{S}$ and each agent $A^i$ obtains an
observation $o^i_t$ recording some partial information about
$s_t$. For instance, $o^i_t$ might be the egocentric visual view of an
agent $A^i$ embedded in some simulated environment. %
From observation $o^i_t$ and history $h^i_{t-1}$, which records prior observations and decisions made by the agent, each agent $A^i$  forms a policy
$\pi^i_t:\cA\to[0,1]$ where $\pi^i_t(a)$  is the probability that agent $A^i$ chooses to take action $a\in\cA$ from a finite set of options $\cA$ at time $t$. 
After the agents execute their respective
actions $(a^1_t,\ldots,a^N_t)$, which we call a \emph{multi-action}, they enter
a new state $s_{t+1}$ and receive individual rewards
$r^1_{t},\ldots,r^N_{t}\in\mathbb{R}$. For more on RL 
see~\cite{SuttonMIT1998,MnihNature2015,MnihEtAlPMLR2016}.

\noindent\textbf{Task definition.} \task is set in the near-photorealistic and physics enabled simulated environment \thor~\cite{ai2thor}. In \task, $N$ agents collaborate to move a lifted object through an indoor environment with the goal of placing this object above a visually distinct target as illustrated in \figref{fig:teaser}. Akin to humans moving large items,  agents must navigate around other furniture and frequently walk in-between obstacles on the floor. %

In \task, each agent at every timestep receives an egocentric observation (a $3\times 84 \times 84$ RGB image) from \thor. In addition, agents are allowed to communicate with other agents at each timestep via a low bandwidth communication channel. Based on their local observation and communication, each agent must take an action from the set $\mathcal{A}$. 
The space of actions $\mathcal{A} = \navactions\cup \mwoactions\cup \moactions\cup \roactions$ available to an agent is comprised
of the 
four single-agent navigational actions $\navactions = \{$\textsc{MoveAhead}, \textsc{RotateLeft}, \textsc{RotateRight}, \textsc{Pass}$\}$ used to move the agent independently, 
four actions $\mwoactions=\{$\textsc{MoveWithObjectX} $\mid X\in \{$\textsc{Ahead, Right, Left, Back}$\}\}$ used to move the lifted object and the agents simultaneously in the same direction,
four actions $\moactions=\{$\textsc{MoveObjectX}$ \mid X\in \{$\textsc{Ahead, Right, Left, Back}$\}\}$ used to move the lifted object while the agents stay in place, 
and a single action used to rotate the lifted object clockwise $\roactions= \{\textsc{RotateObjectRight}\}$. 
We assume that all movement actions for agents and the lifted object result in a displacement of 0.25 meters (similar to~\cite{jain2019CVPRTBONE,habitat19iccv}) and all rotation actions result in a rotation of 90 degrees (counter-)clockwise when viewing the agents from above. %

\begin{figure}[t]
    \centering
    {
        \phantomsubcaption\label{fig:coordination_matrix_furnmove}
        \phantomsubcaption\label{fig:coordination_matrix_furnlift}
    }
    \begin{tabular}{@{\hskip0pt}c@{\hskip15pt}c@{\hskip0pt}}
        \includegraphics[height=5.5cm]{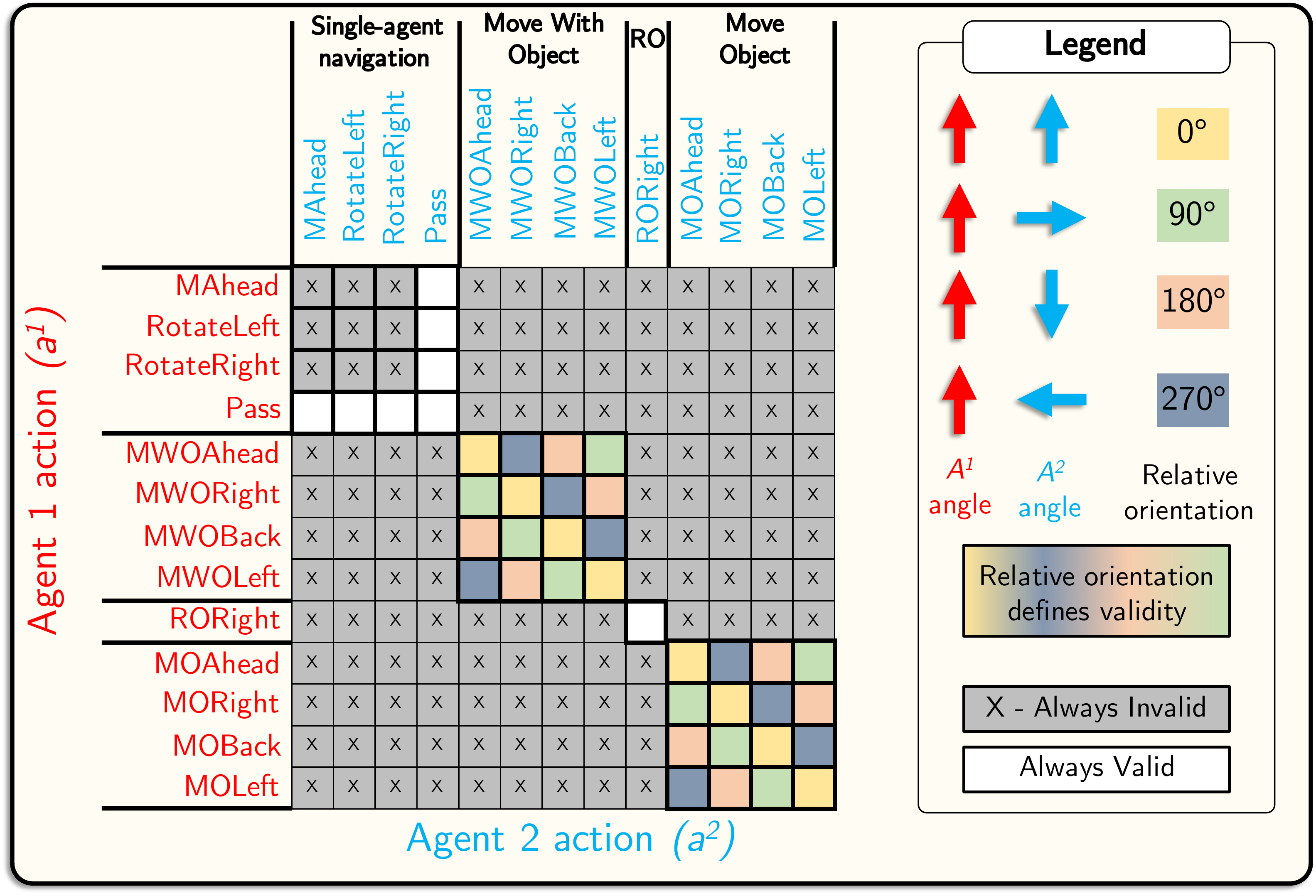} &
        \includegraphics[height=5.5cm]{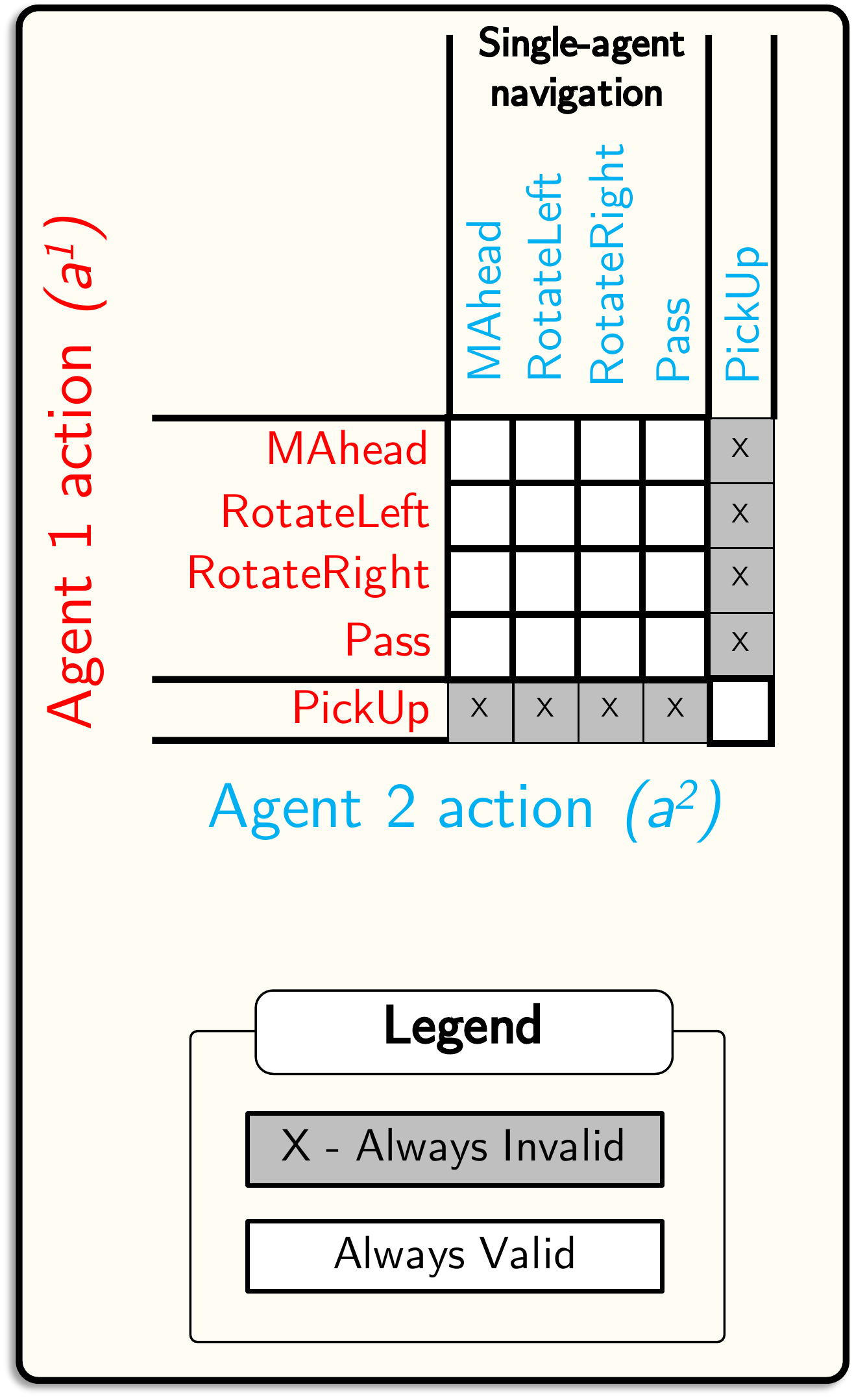}
        \\
        (a) {\task} & (b) {\oldtask}
    \end{tabular}
    \caption{\textbf{Coordination matrix for tasks.} The matrix $S_t$ records the validity of multi-action $(a^1,a^2)$ for different relative orientations of agents \textcolor{red}{$A^1$} \& \textcolor{cyan}{$A^2$}. (a) Overlay of $S_t$ for all four relative orientation of two agents, for {\task}. Notice that only $16/169=9.5\%$ multi-actions are coordinated at any given relative orientation, (b) {\oldtask} where single agent actions are always valid and coordination is needed only for \textsc{PickUp} action, i.e. at least $16/25=64\%$ actions are always valid.}
    \label{fig:coordination_matrix}
\end{figure}

Close and on-going collaboration is required in \task\ due to restrictions on the set of actions which can be successfully completed jointly by all the agents. These restrictions reflect physical constraints: for instance, if two people attempt to move in opposite directions while carrying a heavy object they will either fail to move or drop the object. For two agents, we summarize these restrictions using the \emph{coordination matrix} shown in Fig.~\ref{fig:coordination_matrix_furnmove}. For comparison, we include a similar matrix in \figref{fig:coordination_matrix_furnlift} corresponding to the \oldtask task from~\cite{jain2019CVPRTBONE}. We defer a more detailed discussion of these restrictions to \secref{sec:action-restrictions} of the \appendixorsupp. Generalizing the coordination matrix shown in Fig.~\ref{fig:coordination_matrix_furnmove}, at every timestep $t$ we let $S_t$ be the $\{0,1\}$-valued $|\cA|^N$-dimensional tensor where $(S_t)_{i_1,\ldots,i_N}=1$ if and only if the agents are configured such that multi-action $(a^{i_1},\ldots,a^{i_N})$ satisfies the restrictions detailed in \secref{sec:action-restrictions}. If $(S_t)_{i_1,\ldots,i_N}=1$ we say the actions $(a^{i_1},\ldots,a^{i_N})$ are \emph{coordinated}.

\subsection{Technical challenges} \label{sec:challenges}

As we show in our experiments in \secref{sec:experiments}, standard communication-based models similar to the ones proposed in \cite{jain2019CVPRTBONE} perform rather poorly when trained to complete the \task task. In the following we identify two key challenges that contribute to this poor performance. 

\noindent\textbf{Challenge 1: rank-one joint policies.} In classical multi-agent
settings~\cite{Busoniu2008,Panait2005,LoweNIPS2017}, each agent $A^i$ samples its
action $a^i_t \sim \pi^i_t$ independently of all other agents.
Due to this independent sampling,
at time $t$, the probability of the agents taking multi-action 
$(a^1,\ldots,a^N)$ equals $\prod_{i=1}^N \pi^i_{t}(a^i)$. This means that
the joint probability tensor of all actions at time $t$ can be written as the
rank-one tensor $\Pi_t = \pi^1_t\otimes\cdots \otimes \pi^N_t$.
This rank-one constraint limits the joint policy
that can be executed by the agents, which has real impact.  
\secref{sec:rank-one-challenge-example} considers two agents playing rock-paper-scissors with an adversary:  the rank-one constraint reduces the expected reward achieved by an optimal policy from 0 to -0.657 (minimal reward being -1). Intuitively, a high-rank joint policy is not well approximated by a rank-one probability tensor obtained via independent sampling. 

\noindent\textbf{Challenge 2: exponential failed actions.} 
The number of possible
multi-actions $|\mathcal{A}|^N$ increases exponentially as the number of agents $N$ grows. While this
is not problematic if agents act relatively
independently, it's a significant obstacle when the agents are
\emph{tightly coupled}, \ie, when %
the success of
agent $A^i$'s action $a^i$ is highly dependent on the actions of the other
agents. 
Just consider a randomly initialized policy (the starting point of almost all RL problems): agents stumble upon positive rewards with an extremely low probability which leads to slow learning. We focus on small $N$, nonetheless, the proportion of coordinated action tuples is  small ($9.5\%$ when $N=2$ and $2.1\%$ when $N=3$).

\section{A cordial sync} \label{sec:method}
To address the aforementioned two challenges we develop: (a) a novel action sampling procedure named \textbf{S}ynchronize \textbf{Y}our actio\textbf{N}s \textbf{C}oherently (\method) and (b) an intuitive \& effective multi-agent training loss named the \textbf{C}o\textbf{ordi}n\textbf{a}tion \textbf{L}oss (\loss).

\begin{figure}[t]
    \centering
    \includegraphics[width=\textwidth,trim={0 0 2cm 0},clip]{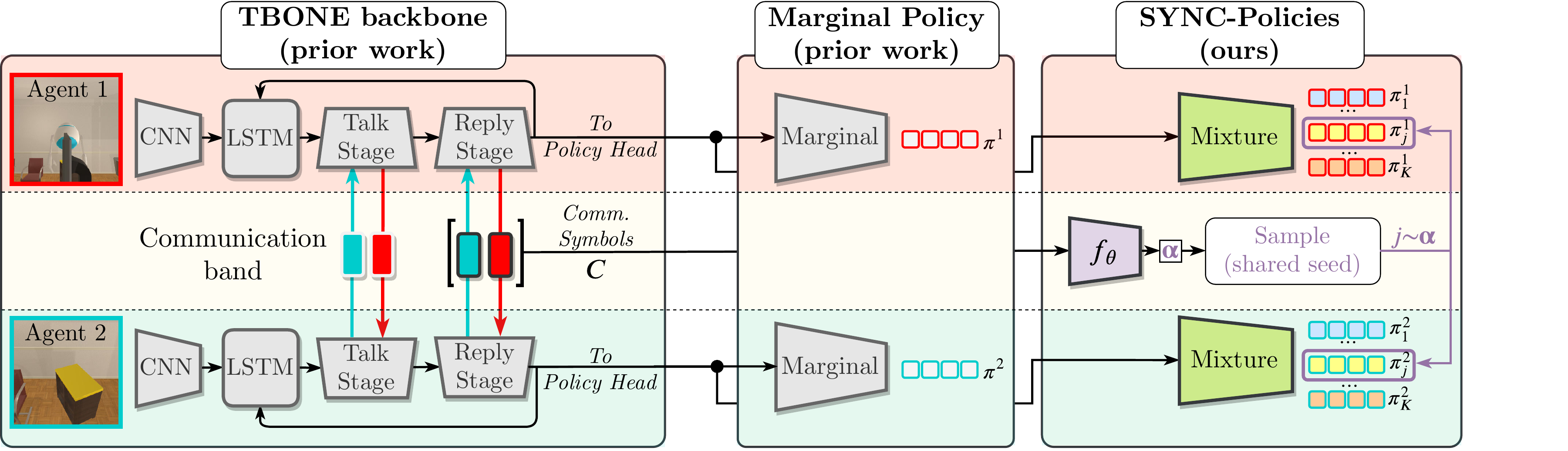}
    \caption{Model overview for 2 communicative agents in the decentralized setting. \textit{Left}: all decentralized methods in this paper have the same TBONE~\cite{jain2019CVPRTBONE} backbone architecture. \textit{Right}: marginal vs \method-policies. With marginal policies, the standard in prior work, each agent constructs its own policy and independently samples from this policy. With \method-policies, agents communicate to construct a distribution $\alpha$ over multiple ``strategies'' which they then sample from using a shared random seed}
    \label{fig:model}
\end{figure}

\noindent\textbf{Addressing challenge 1: \method-policies.} %
For readability, we consider  
$N=2$ agents and illustrates an overview in \figref{fig:model}. The joint probability tensor $\Pi_t$ is hence a matrix of size $|\cA|\times|\cA|$.
Recall our goal: %
using little communication, multiple agents should sample their actions from a high-rank joint policy. This is difficult as (i) little communication means that, except in degenerate cases, no agent can form the full joint policy and (ii) even if all agents had access to the joint policy it is not obvious how to ensure that the decentralized agents will sample a valid coordinated action. %

To achieve this note that, for any rank $m\leq |\cA|$ matrix $L\in\bR^{|\cA|\times |\cA|}$, there are vectors $v_1,w_1,\ldots,v_m,w_m\in\bR^{|\cA|}$ such that
$L = \sum_{j=1}^m v_j\otimes w_j$. Here, $\otimes$ denotes the outer product. 
Also, the \emph{non-negative rank} of a matrix $L\in\bR^{|\cA|\times |\cA|}_{\geq 0}$ equals the smallest integer $s$ such that $L$ can be written as the sum of $s$ non-negative rank-one matrices. 
Furthermore, a non-negative matrix $L\in\bR^{|\cA|\times |\cA|}_{\geq 0}$ has non-negative rank bounded above by $|\cA|$. Since  $\Pi_t$ is a $|\cA|\times |\cA|$ joint probability matrix, \ie, $\Pi_t$ is non-negative and its entries sum to one, it has non-negative rank $m\leq |\cA|$, \ie, there exist non-negative vectors $\alpha\in\bR_{\geq 0}^m$ and
$p_1, q_1,\ldots,p_m,q_m\in\bR_{\geq 0}^{|\cA|}$ whose entries sum to one such that
$\Pi_t=\sum_{j=1}^m\alpha_j \cdot p_j\otimes q_j$. 

We call a sum of the form
$\sum_{j=1}^m\alpha_j \cdot p_j\otimes q_j$ a \emph{mixture-of-marginals}. With this decomposition at hand, randomly sampling action pairs $(a^1,a^2)$ from
$\sum_{j=1}^m\alpha_j \cdot p_j\otimes q_j$ can be interpreted as a
two step process: first sample an index
$j\sim \text{Multinomial}(\alpha)$ and then sample
$a^1\sim \text{Multinomial}(p_j)$ and
$a^2\sim \text{Multinomial}(q_j)$. 

This stage-wise procedure suggests a strategy for sampling actions in a multi-agent setting, which we refer to as
\emph{\method-policies}. Generalizing to an $N$ agent setup, suppose
that agents $(A^i)_{i=1}^N$ have access to a shared random stream of numbers. 
This can be accomplished if all agents
share a random seed or if all agents initially communicate their individual random
seeds and sum them to obtain a shared
seed. Furthermore, suppose that all agents locally store a shared function %
$f_\theta: \mathbb{R}^K\to \Delta_{m-1}$ where $\theta$ are learnable
parameters, $K$ is the dimensionality of all communication
between the agents in a timestep, and $\Delta_{m-1}$ is the
standard $(m-1)$-probability simplex. Finally, at time $t$ suppose
that each agent $A^i$ produces not a single policy $\pi^i_t$ but instead a
collection of policies $\pi^i_{t,1}, \ldots, \pi^i_{t,m}$. Let
$C_t\in\bR^K$ be all communication sent between agents at time $t$. Each agent $A^i$ then samples its action as follows: 
(i) compute
the shared probabilities $\alpha_t = f_\theta(C_t)$, (ii) sample an
index $j\sim \text{Multinomial}(\alpha_t)$ using the shared random
number stream, (iii) sample, independently, an action $a^i$ from the
policy $\pi^i_{t, j}$. Since both
$f_\theta$ and the random number stream are shared, 
the quantities in (i) and (ii) are
equal across all agents despite being computed individually. 
This sampling procedure is equivalent to sampling
from the tensor
$\sum_{j=1}^m \alpha_j \cdot \pi^1_{t, j} \otimes \ldots \otimes \pi^N_{t, j}$ which,
as discussed above, may have rank up to $m$. Intuitively, \method\ enables decentralized agents to have a more expressive joint policy by allowing them to agree upon a strategy by sampling from $\alpha_t$.

\noindent\textbf{Addressing challenge 2: \loss. } %
We encourage agents to rapidly learn to choose coordinated actions via a new loss. In particular,
letting $\Pi_t$ be the joint policy of our agents, we propose the 
\emph{coordination loss} (\loss)
\begin{align}
    \text{CL}_\beta(S_t,\Pi_t) = -\beta \cdot \langle S_t, \log(\Pi_t) \rangle \ / \sum_{1\leq i,j\leq |\cA|} (S_t)_{ij},
    \label{eq:loss}
\end{align}
where $\log$ is applied element-wise, $\langle * , * \rangle$ is the usual Frobenius inner product, and $S_t$ is defined in \secref{sec:task}. Notice that \loss encourages agents to have a near
uniform policy over the actions which are coordinated.
We use this loss to replace the standard entropy encouraging loss in policy gradient  algorithms (\eg, the A3C algorithm~\cite{MnihEtAlPMLR2016}). Similarly to the
parameter for the entropy  loss in A3C, $\beta$ is chosen to be
a small positive constant so as to not overly discourage
learning. 

Note that the coordination loss is less
meaningful when $\Pi_t = \pi^1\otimes \cdots \otimes \pi^N$, \ie, when $\Pi_t$ is rank-one.
For instance, suppose that $S_t$ has
ones along the diagonal, and zeros elsewhere, so that we wish to
encourage the agents to all take the same action. In this case it is
straightforward to show that
$\text{CL}_\beta(S_t,\Pi_t) = -\beta \sum_{i=1}^N \sum_{j=1}^M (1/M)
\log\pi^i_{t}(a^j)$ so that $\text{CL}_\beta(S_t,\Pi_t)$ simply
encourages each agent to have a uniform distribution over its actions and thus actually encourages the agents to place a large amount of probability mass on uncoordinated actions. Indeed, \tabref{tab:cl_study} shows that using \loss without \method leads to poor results.

\section{Models}
\label{sec:baselines}
We study four distinct policy types: \emph{central}, \emph{marginal}, \emph{marginal w/o comm}, and {\itmethod}. \emph{Central} samples actions from a joint policy generated by a central agent with access to observations from all agents. 
While often unrealistic in practice due to communication bottlenecks, \emph{central} serves as an informative baseline. 
\emph{Marginal} follows prior work, \eg,  \cite{jain2019CVPRTBONE}: each agent independently samples its actions from its individual policy after communication. \emph{Marginal w/o comm} is identical to \emph{marginal} but does not permit agents to communicate explicitly (agents may still see each other). %
Finally, {\itmethod} is our newly proposed policy described in \secref{sec:method}. %
For a fair comparison, 
all decentralized agents (\ie, \itmethod, \emph{marginal}, and \emph{marginal w/o comm}), use the same TBONE backbone architecture from \cite{jain2019CVPRTBONE}, see \figref{fig:model}. We have ensured that parameters are fairly balanced so that our proposed \itmethod has close to (and never more) parameters than the \emph{marginal} and \emph{marginal w/o comm} nets. 
Note, we train \emph{central} and {\itmethod} with {\loss}, and the \emph{marginal} and \emph{marginal w/o comm} without it. This choice is mathematically explained in~\secref{sec:method} and empirically validated in~\secref{sec:quantitative}. 

\noindent \textbf{Architecture details:} %
For readability we describe the policy and value net  for the 2 agent setup while noting that it can be trivially extended to any number of agents. 
As noted above, decentralized agents use the  TBONE backbone  from~\cite{jain2019CVPRTBONE}. Our primary architectural novelty extends TBONE to \method-policies. An overview of the TBONE backbone and differences between sampling with \emph{marginal} and \itmethod policies is shown in \figref{fig:model}.

As a brief summary of TBONE, agent $i$ observes at time $t$ inputs $o_t^i$, \ie, %
a $3\times 84 \times 84$ RGB image returned from \thor which represents the $i$-th agent's egocentric view. For each agent, the observation is encoded by a four layer CNN and combined with an agent specific learned embedding (that encodes the ID of that agent) along with the history embedding $h^i_{t-1}$. The resulting vector is fed into a long-short-term-memory (LSTM) \cite{HochreiterNC1997} unit to produce a $512$-dimensional embedding $\tilde{h}^i_t$ corresponding to the $i^\text{th}$ agent.

The agents then undergo two rounds of communication resulting in two final hidden states $h^1_t, h^2_t$ and messages $c^i_{t,j}\in\bR^{16}$, $1\leq i,j\leq 2$ with message $c^i_{t,j}$ being produced by agent $i$ in round $j$ and then sent to the other agent in that round. In \cite{jain2019CVPRTBONE}, the value of the agents' state as well as logits corresponding to the policy of the agents are formed by applying linear functions to $h^1_t, h^2_t$.

We now show how \method can be integrated into TBONE to allow our agents to represent high rank joint distributions over multi-actions (see Fig.~\ref{fig:model}). First each agent computes the logits corresponding to $\alpha_t$. This is done using a $2$-layer MLP applied to the messages sent between the agents, at the second stage. In particular, 
$\alpha_t = {\bf W}_3\ \text{ReLU}({\bf W}_2\ \text{ReLU}({\bf W}_1\ [c^1_{t,2}; c^2_{t,2}] + {\bf b}_1) + {\bf b}_2) + {\bf b}_3$ where ${\bf W}_1\in\bR^{64\times 32}, {\bf W}_2\in\bR^{64\times 64}$,
${\bf W}_3\in\bR^{m\times 64}$, 
${\bf b}_1\in\bR^{32},{\bf b}_2\in\bR^{64}$, and ${\bf b}_3\in\bR^m$ 
are a learnable collection of weight matrices and biases. After computing $\alpha_t$ we compute a collection of policies $\pi^i_{t, 1}, \dots, \pi^i_{t, m}$ for $i\in\{1,2\}$. Each of these policies is computed following the TBONE architecture but using $m-1$ additional, and learnable, linear layers per agent. %

\section{Experiments}
\label{sec:experiments}

\subsection{Experimental setup}
\label{sec:environment}

\noindent\textbf{Simulator.}
We evaluate our models using the \thor  environment~\cite{ai2thor} with several novel upgrades. First, we introduce new methods which permit to (a) randomly initialize the lifted object and agent locations close to each other and looking towards the lifted object, and (b) simultaneously move agents and the lifted object in a given direction with collision detection. 
Secondly, we build a top-down gridworld version of \thor for faster prototyping, that is $16\times$ faster than~\cite{jain2019CVPRTBONE}. For details about framework upgrades, see \secref{sec:extra-training-details} of the \appendixorsupp.

\noindent\textbf{Tasks.}
We compare against baselines on  {\task}, {\gridtask}, and {\oldtask}~\cite{jain2019CVPRTBONE}. {\task} is the novel task introduced in this work (\secref{sec:task}): agents observe egocentric visual views, with field-of-view 90 degrees. In {{\task}-Gridworld} the agents are provided a top-down egocentric 3D tensor as observations. The third dimension of the tensor contains semantic information such as, if the location is navigable by an agent or navigable by the lifted object, or whether the location is occupied by another agent, the lifted object, or the goal object. Hence, {{\task}-Gridworld} agents do not need visual understanding, but face other challenges of the {\task} task -- coordinating actions and planning trajectories. We consider only the harder variant of {\oldtask}, where communication was shown to be most important (`constrained' with no implicit communication in~\cite{jain2019CVPRTBONE}). In {\oldtask}, agents observe egocentric visual views.

\noindent\textbf{Data.}
As in~\cite{jain2019CVPRTBONE}, we train and evaluate on a split of the $30$ living room scenes. As  \task{}  is already quite challenging, we only consider a single piece of lifted furniture (a television) and a single goal object (a TV-stand). %
Twenty rooms are used for training, $5$ for validation, and $5$ for testing. The test scenes have very different lighting conditions, furniture, and layouts. For evaluation our test set includes $1000$ episodes equally distributed over the five scenes. 

\noindent\textbf{Training.} 
For training we augment the A3C algorithm~\cite{MnihEtAlPMLR2016} with \loss. For our studies in the visual domain, we use 45 workers and 8 GPUs. Models take around two days to train. %
For more details about training, including hyperparameter values and the reward structure, see  \secref{sec:extra-training-details} of the \appendixorsupp.

\subsection{Metrics}
\label{sec:metrics}
For completeness, we consider a variety of metrics. We adapt SPL, \ie, Success
weighted by (normalized inverse) Path Length~\cite{anderson2018evaluation}, so that it doesn't require shortest paths but still provides similar semantic information\footnote{For {\task}, each location of the lifted furniture corresponds to $404,480$ states, making shortest path computation intractable (more details in \secref{sec:quant-eval-extra-details} of the \appendixorsupp).}: We define a Manhattan Distance based SPL as $\text{\myspl} = N_\text{ep}^{-1}\sum_{i=1}^{N_\text{ep}}S_i\frac{m_i/d_\text{grid}}{\max(p_i, m_i/d_\text{grid})}$,
where $i$ denotes an index over episodes, $N_\text{ep}$ equals the number of test episodes, and $S_i$ is a binary indicator for success of episode $i$. Also $p_i$ is the number of actions taken per agent, $m_i$ is the Manhattan distance from the lifted object's start location to the goal, and $d_\text{grid}$ is the distance between adjacent grid points, for us $0.25$m. We also report other metrics capturing complementary information. These include mean number of actions in an episode per agent ({\eplen}), success rate ({\srate}), and distance to target at the end of the episode ({\fdist}). 

We also introduce two metrics unique to coordinating actions: {\lrankdist}, the mean total variation distance between $\Pi_{t}$ and its best rank-one approximation, and {\invalidprob}, the average probability mass allotted to uncoordinated actions, \ie, the dot product between $1 - S_t$ and $\Pi_t$.
By definition, {\lrankdist} is zero for the \textit{marginal} model, and higher values indicate divergence from independent marginal sampling. Note that, without measuring \lrankdist we would have no way of knowing if our \method model was actually using the extra expressivity we've afforded it.
Lower {\invalidprob} values imply an improved ability to avoid uncoordination actions as detailed in \secref{sec:task} and~\figref{fig:coordination_matrix}. 

\subsection{Quantitative evaluation}\label{sec:quantitative}
We conduct four  studies: (a) performance of different methods and relative difficulty of the three tasks, (b) effect of number of components on {\method} performance, (c) effect of {\loss} (ablation), and (d) effect of number of agents.

\begin{table}[t]
\caption{Quantitative results on three tasks. $\uparrow$ (or $\downarrow$) indicates that higher (or lower) value of the metric is desirable while $\updownarrow$ denotes that the metric is simply informational and no value is, a priori, better than another. $^\dagger$denotes that a centralized agent serves only as an upper bound to decentralized methods and cannot be fairly compared with. Note that, among decentralized agents, our \method model has the best metric values across all reported metrics (\textbf{bolded} values). Values are {\hl{highlighted in green}} if their 95\% confidence interval has no overlap with the confidence intervals of other values
} \label{tab:quant}
\setlength{\tabcolsep}{6pt}
\centering
\resizebox{0.9\linewidth}{!}{%
\begin{tabular}{ccccccc}
\hline
\rowcolor[HTML]{EFEFEF} 
\textbf{Methods} &
\textbf{MD-SPL} $\uparrow$ &
\textbf{Success} $\uparrow$&
\textbf{Ep len} $\downarrow$&
\textbf{\begin{tabular}[c]{@{}c@{}}Final \\ dist\end{tabular}} $\downarrow$& \textbf{\begin{tabular}[c]{@{}c@{}}Invalid \\ prob.\end{tabular}} $\downarrow$&
\textbf{TVD} $\updownarrow$\\ 
\hline
\multicolumn{7}{c}{{\task} (ours)} \\ \hline
Marginal w/o comm~\cite{jain2019CVPRTBONE} & 0.032 & 0.164 & 224.1 & 2.143 & 0.815 & 0 \\
Marginal~\cite{jain2019CVPRTBONE} & 0.064 & 0.328 & 194.6 & 1.828 & 0.647 & 0 \\
{\method} & \hl{\textbf{0.114}} & \hl{\textbf{0.587}} & \hl{\textbf{153.5}} & \hl{\textbf{1.153}} & \hl{\textbf{0.31}} & 0.474 \\
\hdashline
Central$^\dagger$ & 0.161 & 0.648 & 139.8 & 0.903 & 0.075 & 0.543 \\ \hline

\multicolumn{7}{c}{{\gridtask} (ours)} \\ \hline
Marginal w/o comm~\cite{jain2019CVPRTBONE} & 0.111 & 0.484 & 172.6 & 1.525 & 0.73 & 0 \\
Marginal~\cite{jain2019CVPRTBONE} & 0.218 & 0.694 & 120.1 & 0.960 & 0.399 & 0 \\
{\method} & \textbf{0.228} & \hl{\textbf{0.762}} & \textbf{110.4} & \hl{\textbf{0.711}} & \hl{\textbf{0.275}} & 0.429 \\
\hdashline
Central$^\dagger$ & 0.323 & 0.818 & 87.7 & 0.611 & 0.039 & 0.347 \\ \hline

\multicolumn{7}{c}{{\gridtask}-3Agents (ours)} \\ \hline
Marginal~\cite{jain2019CVPRTBONE} & 0 & 0 & 250.0 & 3.564 & 0.823 & 0\\
{\method} & \hl{\textbf{0.152}} & \hl{\textbf{0.578}} & \hl{\textbf{149.1}} & \hl{\textbf{1.05}} & \hl{\textbf{0.181}} & 0.514\\
\hdashline
Central$^\dagger$& 0.066 & 0.352 & 195.4 & 1.522 & 0.138 & 0.521\\
\hline
\end{tabular}
}
\vspace{-4mm}
\end{table}

\begin{table}[t]
\caption{Quantitative results on the \oldtask task. For legend, see \tabref{tab:quant}
} \label{tab:quant-furnlift}
\setlength{\tabcolsep}{6pt}
\centering
\resizebox{\linewidth}{!}{
\begin{tabular}{ccccccccc}
\hline
\rowcolor[HTML]{EFEFEF} 
\textbf{Methods} &
\textbf{MD-SPL} $\uparrow$ &
\textbf{Success} $\uparrow$&
\textbf{Ep len} $\downarrow$&
\textbf{\begin{tabular}[c]{@{}c@{}}Final \\ dist\end{tabular}} $\downarrow$& 
\textbf{\begin{tabular}[c]{@{}c@{}}Invalid \\ prob.\end{tabular}} $\downarrow$&
\textbf{TVD} $\updownarrow$&
\textbf{\begin{tabular}[c]{@{}c@{}}Failed \\pickups\end{tabular}} $\downarrow$& 
\textbf{\begin{tabular}[c]{@{}c@{}}Missed \\pickups\end{tabular}} $\downarrow$\\
\hline
\multicolumn{9}{c}{{\oldtask}~\cite{jain2019CVPRTBONE} (`constrained' setting with no implicit communication)} \\ \hline
Marginal w/o comm~\cite{jain2019CVPRTBONE} & 0.029 & 0.15 & 229.5 & 2.455 & 0.11 & 0 & 25.219 & 6.501 \\
Marginal~\cite{jain2019CVPRTBONE} & \textbf{0.145} & \textbf{0.449} & \textbf{174.1} & 2.259 & 0.042 & 0 & 8.933 & 1.426 \\
{\method} & 0.139 & 0.423 & 176.9 & \textbf{2.228} & \hl{\textbf{0}} & 0.027 & \hl{\textbf{4.873}} & \textbf{1.048} \\
\hdashline
Central$^\dagger$ & 0.145 & 0.453 & 172.3 & 2.331 & 0 & 0.059 & 5.145 & 0.639 \\
\hline
\end{tabular}
}
\end{table}

\begin{figure}[t]
    \centering
    {
        \phantomsubcaption\label{fig:tb-graphs-furnmove}
        \phantomsubcaption\label{fig:tb-graphs-grid-furnmove}
        \phantomsubcaption\label{fig:tb-graphs-furnlift}
    }
    \begin{tabular}{ccc}
        \includegraphics[width=0.33\linewidth]{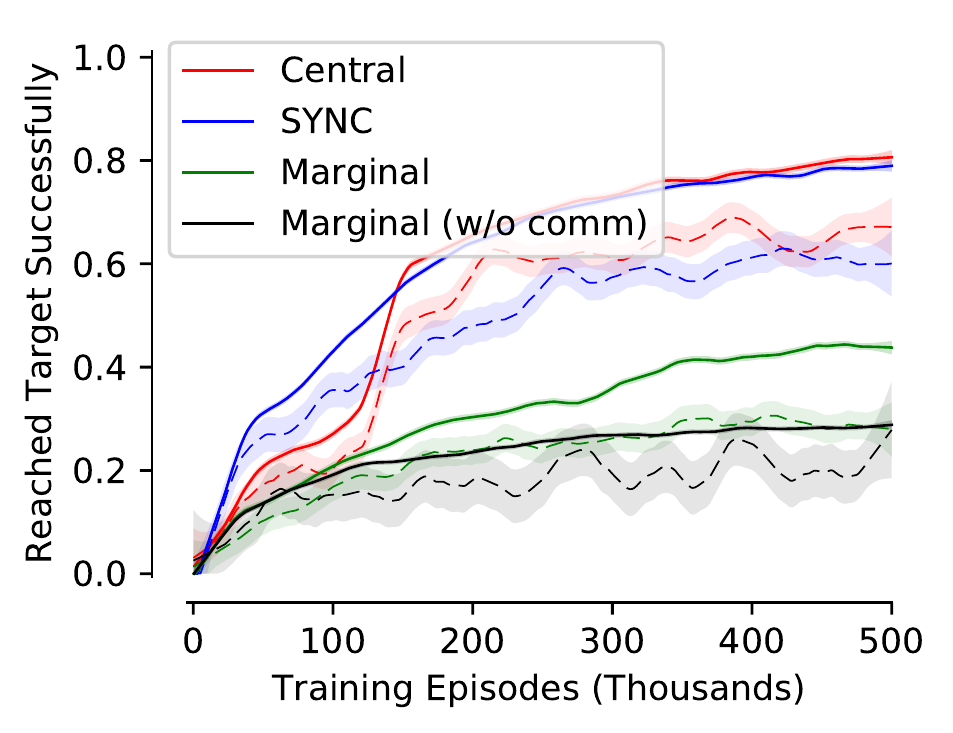} &
        \includegraphics[width=0.33\linewidth]{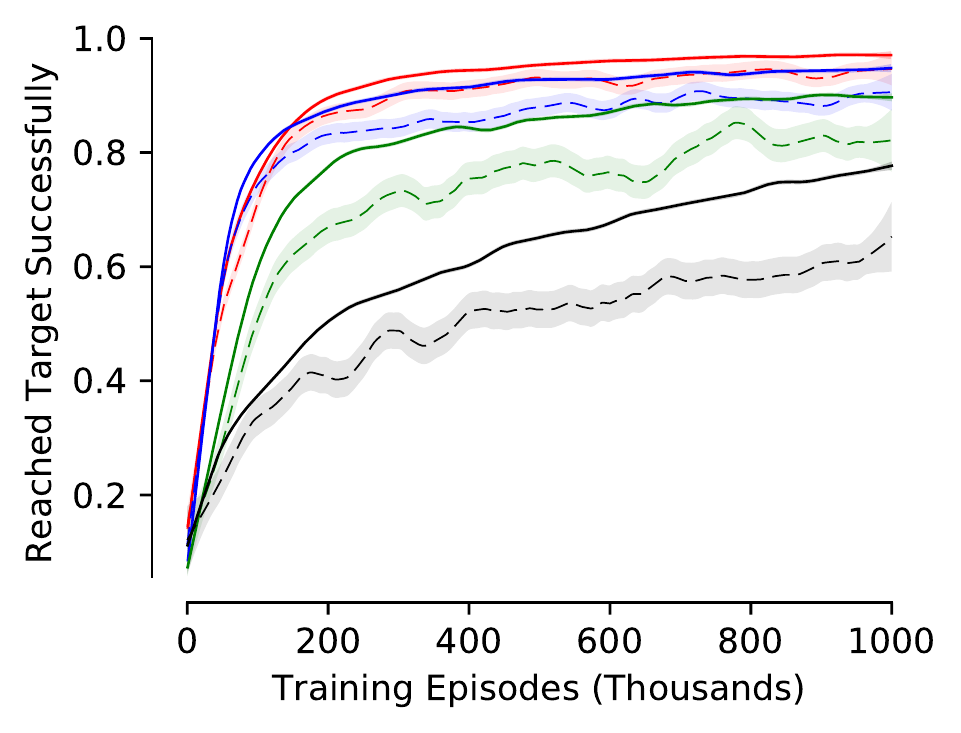} &
        \includegraphics[width=0.33\linewidth]{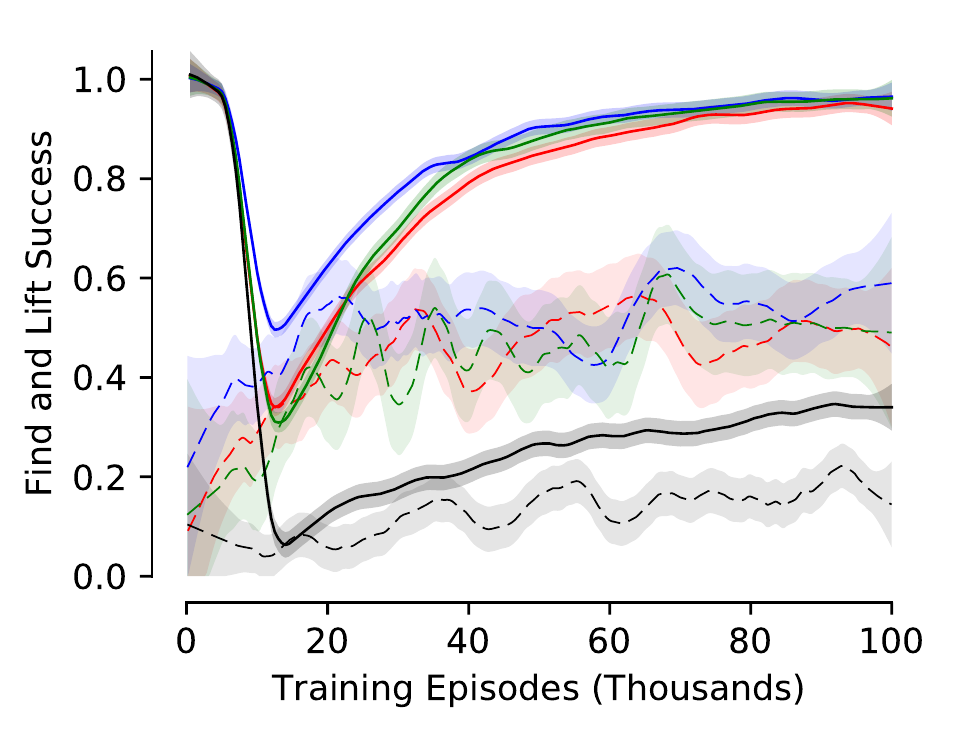} \\
        (a) {\task} &
        (b) {\gridtask} &
        (c) {\oldtask}
    \end{tabular}
    \caption{\textbf{Success rate during training.} Train (solid lines) and validation (dashed lines) performance of our agents for {\task}, {\gridtask}, and {\oldtask}. $95$\% confidence bars are included. For additional plots, see \secref{sec:quant-eval-extra-details} of the {\appendixorsupp}
    }
    \label{fig:tb-graphs}
\vspace{-4mm}
\end{figure}

\noindent\textbf{Comparing methods and tasks.} We compare models detailed in \secref{sec:baselines} on tasks of varying difficulty, report metrics in  \tabref{tab:quant}, and show the progress of metrics over training episodes in \figref{fig:tb-graphs}. In our \task experiments, \itmethod performs better than the best performing method of~\cite{jain2019CVPRTBONE} (\ie, \textit{marginal}) on all metrics. Success rate increases by $25.9$\% and $6.8$\% absolute percentage points on \task and \gridtask respectively. Importantly, \method is significantly better at allowing agents to coordinate their actions: for \task, the joint policy of \method assigns, on average, $0.31$ probability mass to invalid actions pairs while the \textit{marginal} and \textit{marginal w/o comm} models assign $0.647$ and $0.815$ probability mass to invalid action pairs. Additionally, \method goes beyond rank-one \textit{marginal} methods by capturing a more expressive joint policy using the mixture of marginals. This is evidenced by the high TVD of $0.474$ \vs $0$ for \textit{marginal}. In \gridtask, oracle-perception of a 2D gridworld helps raise performance of all methods, though the trends are similar. \tabref{tab:quant-furnlift} shows similar trends for  \oldtask but, perhaps surprisingly, the \srate of \itmethod is somewhat lower than the \textit{marginal} model (2.6\% lower, within statistical error). As is emphasized in \cite{jain2019CVPRTBONE} however, \srate alone is a poor measure of model performance: equally important are the \textit{failed pickups} and \textit{missed pickups} metrics (for details, see \secref{sec:quant-eval-extra-details} of the \appendixorsupp). For these metrics, \itmethod outperforms the \textit{marginal} model. That \itmethod does not completely outperform \textit{marginal} in \oldtask is intuitive, as \oldtask does not require continuous close coordination the benefits of \itmethod are less pronounced.

While the difficulty of a task is hard to quantify, we will consider the relative test-set metrics of agents on various tasks as an informative proxy. Replacing the complex egocentric vision in \task with the semantic 2D gridworld in \gridtask, we see that all agents show large gains in \srate and \myspl, suggesting that \gridtask is a dramatic simplification of \task. Comparing \task to \oldtask is particularly interesting. The \myspl and \srate metrics for the \textit{central} agent do not provide a clear indication regarding task difficulty amongst the two. However, notice the much higher \lrankdist for the \textit{central} agent for \task and the superior \myspl and \srate of the \textit{Marginal} agents for \oldtask. These numbers clearly indicate that \task requires more coordination and additional expressivity of the joint distribution than \oldtask.

\begin{table}[t]
\caption{Effect of number of mixture components $m$ on {\itmethod}'s performance (in {\task}). Generally, larger $m$ means larger \lrankdist values and better performance.
} \label{tab:mixture}
\centering
\setlength{\tabcolsep}{6pt}
\resizebox{0.8\linewidth}{!}{%
\begin{tabular}{c@{\hskip6pt}c@{\hskip6pt}c@{\hskip6pt}c@{\hskip6pt}c@{\hskip6pt}c@{\hskip6pt}c} 
\hline
\rowcolor[HTML]{EFEFEF} 
\textbf{$K$ in {\method}}  &
\textbf{MD-SPL} $\uparrow$ &
\textbf{Success} $\uparrow$&
\textbf{Ep len} $\downarrow$&
\textbf{\begin{tabular}[c]{@{}c@{}}Final \\ dist\end{tabular}} $\downarrow$& \textbf{\begin{tabular}[c]{@{}c@{}}Invalid \\ prob.\end{tabular}} $\downarrow$&
\textbf{TVD} $\updownarrow$\\ \hline
\multicolumn{7}{c}{\task} \\ \hline
1 component & 0.064 & 0.328 & 194.6 & 1.828 & 0.647 & 0 \\
2 components & 0.084 & 0.502 & 175.5 & 1.227 & \textbf{0.308} & 0.206 \\
4 components & 0.114 & 0.569 & 154.1 & \textbf{1.078} & 0.339 & 0.421 \\
13 components & \textbf{0.114} & \textbf{0.587} & \textbf{153.5} & 1.153 & 0.31 & 0.474 \\
\hline
\end{tabular}%
}
\end{table}

\noindent\textbf{Effect of number of mixture components in \method.} Recall (\secref{sec:method}) that the number of mixture components $m$  in \method is a hyperparameter controlling the maximal rank of the joint policy.   \itmethod with $m=1$  is equivalent to  \textit{marginal}. In \tabref{tab:mixture} we see $TVD$ increase from $0.206$ to $0.474$ when increasing $m$  from 2 to 13. This suggests that   \method  learns to use the additional expressivity. Moreover, we see that this increased expressivity results in better performance. A success rate jump of $17.4$\%  from $m=1$ to $m=2$ demonstrates that substantial benefits are obtained by even small increases in  expressitivity.  Moreover with more components, \ie,  $m=4$ \&  $m=13$ we obtain more improvements. Notice however that there are diminishing returns, the $m=4$ model performs nearly as well as the $m=13$ model. This suggests a trade-off between the benefits of expressivity and the increasing complexities in optimization. 

\noindent\textbf{Effect of \loss.} In~\tabref{tab:cl_study} we quantify the effect of \loss. Note the $9.9\%$ improvement in success rate when adding \loss to \itmethod. This is accompanied by a drop in \textit{Invalid prob.} from $0.47$ to $0.31$, which signifies better coordination of actions. Similar improvements are seen for the \textit{central} model. In `Challenge 2' (\secref{sec:method}) we mathematically laid out why \textit{marginal} models gain little from \loss. We substantiate this empirically with a 22.9\% drop in success rate when training the \textit{marginal} model with \loss.

\noindent\textbf{Effect of more agents.} The final three rows of \tabref{tab:quant} show the test-set performance of \itmethod, \textit{marginal}, and \textit{central} models trained to accomplish a 3-agent variant of our \gridtask task. In this task the \textit{marginal} model fails to train at all, achieving a 0\% success rate. \itmethod, on the other hand, successfully completes the task 57.8\% of the time. Notice that \itmethod's success rate drops by nearly 20 percentage points when moving from the 2- to the 3-agent variant of the task: clearly increasing the number of agents substantially increases the task's difficult. Surprisingly, the \textit{central} model performs worse than \itmethod in this setting. A discussion of this phenomenon is deferred to \secref{sec:quant-eval-extra-details} of the \appendixorsupp.

\begin{table}[t]
\caption{Ablation study of coordination loss on \textit{marginal}~\cite{jain2019CVPRTBONE}, \itmethod, and \textit{central} methods. \textit{Marginal} performs better without {\loss} whereas {\itmethod} and \textit{central} show improvement with {\loss} added to overall loss value.
$^\dagger$denotes that a centralized agent serve only as an upper bound to decentralized methods.
}
\label{tab:cl_study}
\centering
\setlength{\tabcolsep}{4pt}
\resizebox{0.9\linewidth}{!}{%
\begin{tabular}{cccccccc} 
\hline\rowcolor[HTML]{EFEFEF} 
\textbf{Method}  &
\textbf{{\loss}} &
\textbf{MD-SPL} $\uparrow$ &
\textbf{Success} $\uparrow$&
\textbf{Ep len} $\downarrow$&
\textbf{\begin{tabular}[c]{@{}c@{}}Final \\ dist\end{tabular}} $\downarrow$& \textbf{\begin{tabular}[c]{@{}c@{}}Invalid \\ prob.\end{tabular}} $\downarrow$&
\textbf{TVD} $\updownarrow$\\ \hline
\multicolumn{8}{c}{ \textit{\task} } \\ 
\hline
Marginal & {\color{red} \ding{55}} &  0.064 & 0.328 & 194.6 & 1.828 & 0.647 & 0\\
Marginal & {\color{green} \ding{51}} & 0.015 & 0.099 & 236.9 & 2.134 & 0.492 & 0\\
\method & {\color{red} \ding{55}} &  0.091 & 0.488 & 170.3 & 1.458 & 0.47 & 0.36\\
\method & {\color{green} \ding{51}} &  \textbf{0.114} & \textbf{0.587} & \textbf{153.5} & \textbf{1.153} & \textbf{0.31} & 0.474\\
\hdashline
Central$^\dagger$ & {\color{red} \ding{55}} &  0.14 & 0.609 & 146.9 & 1.018 & 0.155 & 0.6245\\
Central$^\dagger$ & {\color{green} \ding{51}} &  0.161 & 0.648 & 139.8 & 0.903 & 0.075 & 0.543\\
\hline
\end{tabular}
}
\end{table}

\subsection{Qualitative evaluation}
\label{sec:qualitative}
We present three qualitative results on \task: joint policy summaries, analysis of learnt communication, and visualizations of agent trajectories. %

\begin{figure}[t]
    \centering
    {
        \phantomsubcaption\label{fig:matrices_central}
        \phantomsubcaption\label{fig:matrices_sync}
        \phantomsubcaption\label{fig:matrices_marginal}
        \phantomsubcaption\label{fig:matrices_communication_analysis}
    }
    \includegraphics[width=1.0\linewidth]{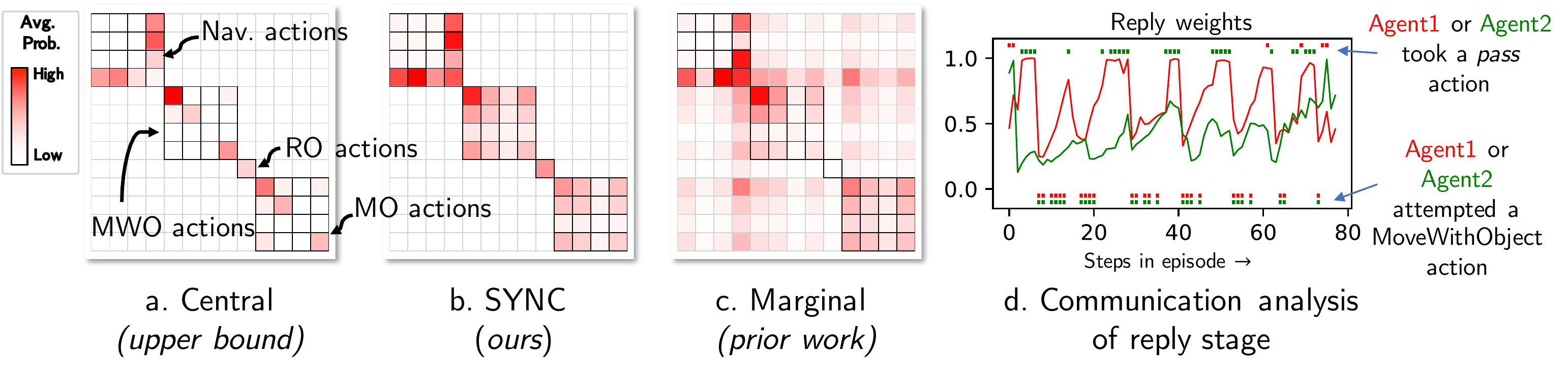}
    \caption{\textbf{Qualitative results.} (a,b,c) joint policy summary ($\Pi_t$ averaged over steps in test episodes in {\task}) and (d) communication analysis. 
    }
    \label{fig:matrices}
\end{figure}

\noindent\textbf{Joint policy summaries.} In~\figref{fig:matrices} we show summaries of the joint policy captured by the \textit{central}, \itmethod, and \textit{marginal} models. These matrices average over action steps in the test-set episodes for \task. Other tasks show similar trends, see \secref{sec:qualitative-extra-details} of the \appendixorsupp. In~\figref{fig:matrices_central}, the sub-matrices corresponding to $\mathcal{A}^{\textsc{MWO}}$ and $\mathcal{A}^{\textsc{MO}}$ are diagonal-dominant, indicating that agents are looking in the same direction ($0\degree$ relative orientation in~\figref{fig:coordination_matrix}). Also note the high probability associated to (\textsc{Pass}, \textsc{RotateX}) and (\textsc{RotateX}, \textsc{Pass}), within the $\mathcal{A}^{\text{NAV}}$ block. Together, this means that the \textit{central} method learns to coordinate single-agent navigational actions to rotate one of the agents (while the other holds the TV by executing \textsc{Pass}) until both face the same direction. They then execute the same action from $\mathcal{A}^{\textsc{MO}}$ ($\mathcal{A}^{\textsc{MWO}}$) to move the lifted object. 
Comparing \figref{fig:matrices_sync}~\vs~\figref{fig:matrices_marginal}, shows the effect of \loss. Recall that the \textit{marginal} model doesn't support {\loss} and thus suffers by assigning probability to invalid action pairs (color outside the block-diagonal submatrices). Also note the banded nature of \figref{fig:matrices_marginal} resulting from its construction as an outer product of marginals.

\noindent\textbf{Communication analysis.} A qualitative discussion of communication follows. Agent are colored red and green. We defer a quantitative treatment to \secref{sec:qualitative-extra-details} of the \appendixorsupp. %
As we apply {\method} on the TBONE backbone introduced by Jain \etal~\cite{jain2019CVPRTBONE}, we use similar tools to understand the  communication emerging %
with {\method} policy heads. In line with~\cite{jain2019CVPRTBONE}, we plot the weight assigned by each agent to the first communication symbol in the reply stage.~\figref{fig:matrices_communication_analysis} strongly suggests that the reply stage is directly used by the agents to coordinate the modality of actions they intend to take. In particular, note that a large weight being assigned to the first reply symbol is consistently associated with the other agent taking a \textsc{Pass} action. %
Similarly, we see that %
small reply weights coincide with agents taking a \textsc{MoveWithObject} action.  %
The talk weights' interpretation is intertwined with the reply weights, and is deferred to \secref{sec:qualitative-extra-details} of the {\appendixorsupp}. 

\noindent\textbf{Agent trajectories.} Our supplementary video includes examples of policy roll-outs. These clips include both agents' egocentric views and a top-down trajectory visualization. This enables direct comparisons of \emph{marginal} and {\method} on the same test episode. We also allow for hearing patterns in agents' communication: we convert scalar weights (associated with reply symbols) to audio. 

\section{Conclusion} \label{sec:conclusion}

We introduce \task, a collaborative, visual, multi-agent task requiring close coordination between agents and develop novel methods that allow for moving beyond existing marginal action sampling procedures, these methods lead to large gains across a diverse suite of metrics.

\noindent\textbf{Acknowledgements:} 
This material is based upon work supported in part by the National Science Foundation under Grants No.\ 1563727, 1718221, 1637479, 165205, 1703166, Samsung, 3M, Sloan Fellowship, NVIDIA Artificial Intelligence Lab, Allen Institute for AI, Amazon, and AWS Research Awards. UJ is thankful to Thomas \& Stacey Siebel Foundation for Siebel Scholars Award. We thank Mitchell Wortsman and Kuo-Hao Zeng for their insightful suggestions on how to clarify and structure this work.

\bibliographystyle{splncs04}
\bibliography{egbib.bib}

\clearpage
\appendix
\section{Supplementary Material
} \label{sec:supp}

This supplementary material provides:

\begin{itemize}
    \item[\ref{sec:action-restrictions}] The conditions for a collection of actions to be considered \textit{coordinated}. 
    \item[\ref{sec:rank-one-challenge-example}] An example showing that standard independent multi-agent action sampling makes it impossible to, even in principle, obtain an optimal joint policy.
    \item[\ref{sec:extra-training-details}] Training details including hyperparameter choices, hardware configurations, and reward structure. We also discuss our upgrades to \thor.
    \item[\ref{sec:quant-eval-extra-details}] Additional discussion, tables, and plots regarding our quantitative results. 
    \item[\ref{sec:qualitative-extra-details}] Additional discussion, tables, and plots of our qualitative results including a description of our supplementary video as well as an in-depth quantitative evaluation of communication learned by our agents. 
\end{itemize}

\subsection{Action restrictions} 
\label{sec:action-restrictions}

We now comprehensively describe the restrictions defining when actions taken by agents are globally consistent with one another. In the following we will, for readability, focus on the two agent setting. All conditions defined here easily generalize to any number of agents. Recall the sets $\navactions, \mwoactions, \moactions$, and $\roactions$ defined in \secref{sec:task}. We call these sets the \emph{modalities of action}. Two actions $a^1,a^2\in\cA$ are said to be of the same modality if they both are an element of the same modality of action. Let $a^1$ and $a^2$ be the actions chosen by the
two agents. Below we describe the conditions  when $a^1$ and $a^2$ are considered \emph{coordinated}. If the agents' actions are uncoordinated, both actions fail and
no action is taken for time $t$. These conditions are summarized in \figref{fig:coordination_matrix_furnmove}.

\noindent\textbf{Same action modality.} A first necessary, but not
sufficient, condition for successful coordination is that the agents
agree on the modality of action to perform. Namely that both $a^1$ and
$a^2$ are of the same action modality. Notice the block diagonal structure  in \figref{fig:coordination_matrix_furnmove}.

\noindent\textbf{No independent movement.} Our second condition models
the intuitive expectation that if one agent wishes to reposition
itself by performing a single-agent navigational action, the other
agent must remain stationary. %
Thus, if $a^1,a^2\in\navactions$, then
$(a^1,a^2)$ are coordinated if and only if one of $a^1$ or $a^2$ is a \textsc{Pass} action. The $\{1,2,3,4\}^2$ entries of the matrix in \figref{fig:coordination_matrix_furnmove} show coordinated pairs of single-agent navigational actions.

\noindent\textbf{Orientation synchronized object movement.} Suppose that 
both agents wish to move (with) the object in a direction so that
$a^1,a^2\in\mwoactions$ or
$a^1,a^2\in\moactions$. Note that, as actions are
taken from an egocentric perspective, it is possible, for example,
that moving ahead from one agent's perspective is the same as moving
left from the other's. This condition requires that the direction specified
by both of the agents is consistent globally. Hence $a^1,a^2$ are
coordinated if and only if the direction specified by both
actions is the same in a global reference frame. For example, if both agents
are facing the same direction this condition requires that
$a^1=a^2$ while if the second agent is rotated 90 degrees
clockwise from the first agent then $a^1=\textsc{MoveObjectAhead}$ will
be coordinated if and only if $a^2=\textsc{MoveObjectLeft}$. See the multicolored 4$\times$4 blocks in \figref{fig:coordination_matrix_furnmove}.

\noindent\textbf{Simultaneous object rotation.} For the lifted 
object to be rotated, both agents must rotate it in the same direction in a global reference frame. As we only allow the
agents to rotate the object in a single direction (clockwise) this
means that %
$a^1=\textsc{RotateObjectRight}$ %
 requires 
$a^2=a^1$. See the (9, 9) entry of the matrix in \figref{fig:coordination_matrix_furnmove}.

While a pair of uncoordinated actions are always unsuccessful, it need not be true that a pair of coordinated actions is successful. 
A pair of coordinated actions will be unsuccessful in two cases: performing the action pair would result in (a) an agent, or the
lifted object, colliding with one another or another object in the
scene; or (b) an agent moving to a position more than 0.76m from the lifted object. Here (a) enforces the physical constraints of the environment while (b) makes the
intuitive requirement that an agent has a finite reach and cannot lift an object when being far away.

\subsection{Challenge 1 (rank-one joint policies) example}
\label{sec:rank-one-challenge-example}
We now illustrate how requiring two agents to independently sample actions from  marginal policies can result in failing to capture an optimal, high-rank, joint policy. 

Consider two agents $A^1$ and $A^2$ who must work together to play
rock-paper-scissors (RPS) against some adversary $E$. In particular, our game takes place in a single timestep where each agent $A^i$, after perhaps communicating with the other agent, must choose some action $a^i\in\cA=\{R, P, S\}$. During this time the adversary also chooses some action $a^E \in \cA$. Now, in our game, the pair of agents $A^1,A^2$ lose if they choose different actions (\ie, $a^1\not=a^2$), tie with the adversary if all players choose the same action (\ie, $a^1=a^2=a^E$), and finally win or lose if they jointly choose an action that beats or losses against the adversary's choice following the normal rules of RPS (\ie, win if $(a^1,a^2,a^E)\in\{(R,R,S),$ $(P,P,R),$ $(S,S,P)\}$, lose if $(a^1,a^2,a^E)\in\{(S,S,R),$ $(R,R,P),$ $(P,P,S)\}$).

Moreover, we consider the challenging setting where $A^1,A^2$
communicate in the open so that the adversary can view their joint policy $\Pi$
before choosing the action it wishes to take. Notice that we've dropped the $t$ subscript on $\Pi$ as there is only a single timestep. Finally, we treat this game as zero-sum so that our agents obtain a reward of 1 for victory, 0 for a tie, and -1 for a loss. We refer to the optimal joint policy as $\Pi^*$. If the agents operate in a decentralized manner using their own (single) marginal policies, their effective rank-one joint policy equals $\Pi = \pi^1\otimes\pi^2$.\\
\noindent\textbf{Optimal joint policy:} It is well known, and easy to show, that
the optimal joint policy equals $\Pi^* = I_{3} / 3$, where $I_{3}$ is the identity matrix of size $3$. Hence, the agents take
multi-action ($R, R$), ($P,P$), or ($S,S$)
with equal probability obtaining an expected reward of zero.\\
\noindent\textbf{Optimal rank-one joint policy:} $\Pi^*$ (the optimal joint policy) is of rank three and thus cannot be captured by $\Pi$ (an outer product of marginals). Instead, brute-force symbolic minimization, using Mathematica \cite{Mathematica}, shows that an optimal strategy for $A^1$ and $A^2$ is to let $\pi^1=\pi^2$ with
\begin{align}
    \pi^1(R) &= 2-\sqrt{2} \approx 0.586, \\
    \pi^1(P) &= 0, \text{ and }\\
    \pi^1(S) &= 1 - \pi^1(R) \approx 0.414.
\end{align}
The expected reward from this strategy is $5 - 4\sqrt{2}\approx- .657$, far
less than the optimal expected reward of $0$.

\subsection{Training details}
\label{sec:extra-training-details}

\subsubsection{Centralized agent.} 
\begin{figure}[t]
    \centering
    \includegraphics[trim={0 0 2cm 0},clip,width=\linewidth]{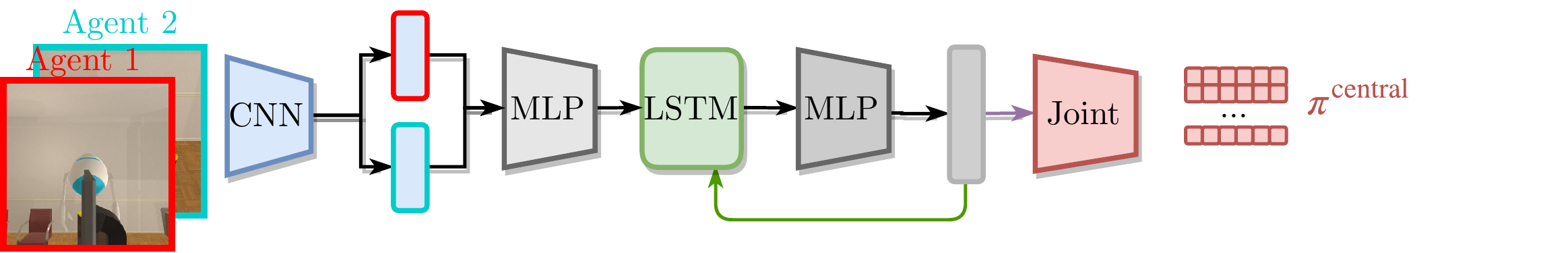}
    \caption{\textbf{Central model architecture.} The central backbone observes the aggregate of all agents' observations. Moreover, the actor in the central model explicitly captures the joint policy distribution.}
    \label{fig:central_model}
\end{figure}

\figref{fig:central_model} provides an overview of the architecture of the centralized agent. The final joint policy is constructed using a single linear layer applied to a hidden state. As this architecture varies slightly when changing the number of agents and the environment (\ie, \thor or our gridworld variant of \thor) we direct anyone interested in exact replication to our codebase.

\subsubsection{AI2-THOR upgrades.}
\begin{figure}[t]
    \centering
    \includegraphics[width=\textwidth]{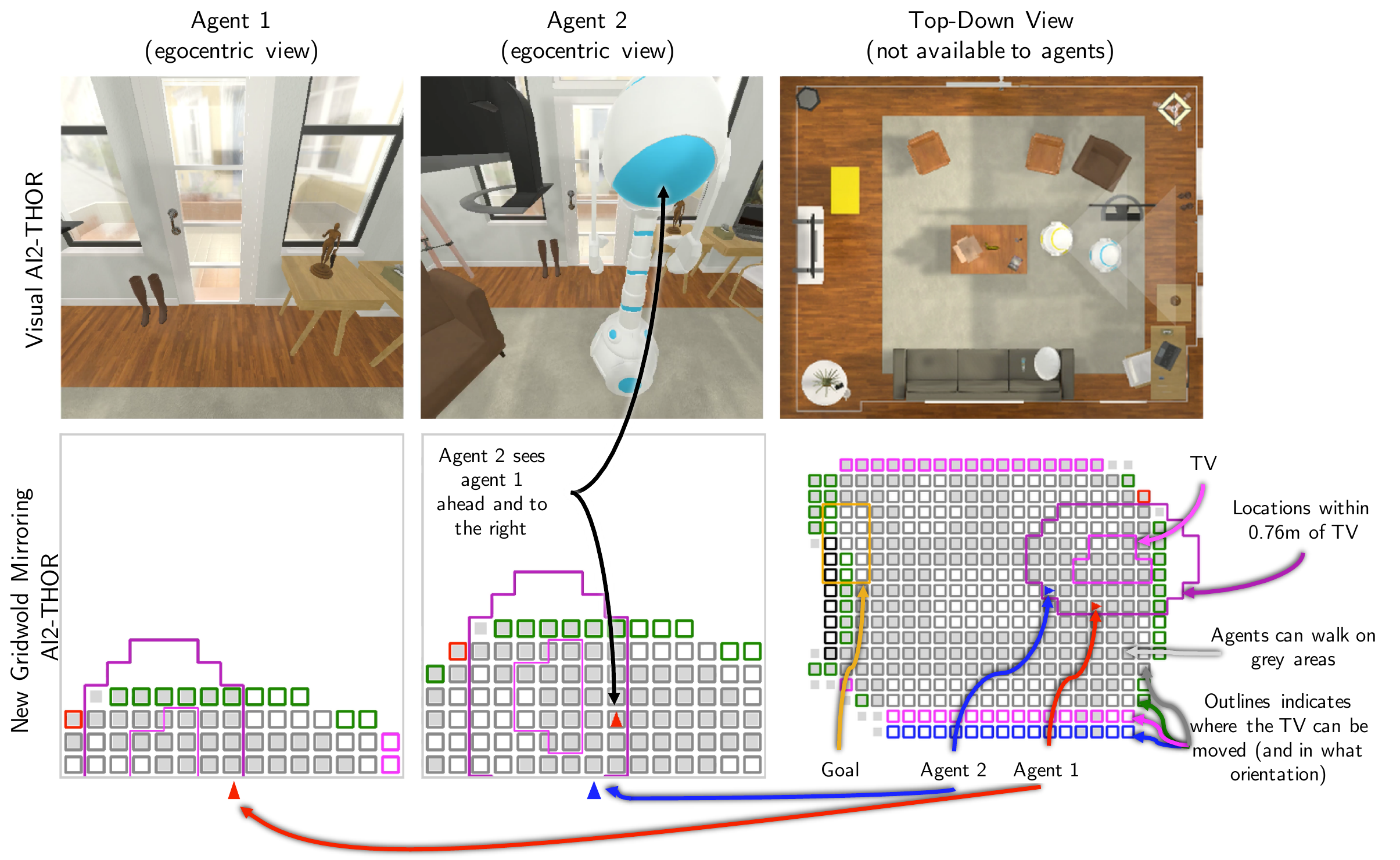}
    \caption{\textbf{Directly comparing visual \thor with our gridworld.} The same scene with identical agent, TV, and TV-stand, positions in \thor and our gridworld mirroring \thor. Gridworld agents receive clean, task-relevant, information directly from the environment while, in \thor, agents must infer this information from complex images.}
    \label{fig:grid_vision_thor_comparison}
\end{figure}

As we described in \secref{sec:experiments} we have made several upgrades to \thor in order to make it possible to run our \task task. These upgrades are described below. \\
\noindent\textbf{Implementing \task methods in \thor's Unity codebase.} The \thor simulator has been built using C\# in Unity. While multi-agent support exists in \thor, our \task task required implementing a collection of new methods to support randomly initializing our task and moving agents in tandem with the lifted object. Initialization is accomplished by a randomized search procedure that first finds locations in which the lifted television can be placed and then determines if the agents can be situated around the lifted object so that they are sufficiently close to the lifted object and looking at it. Implementing the joint movement actions (recall $\mwoactions$) required checking that all agents and objects can be moved along straight-line paths without encountering collisions. \\
\noindent\textbf{Top-down Gridworld Mirroring \thor.} To enable fast prototyping and comparisons between differing input modalities, we built an efficient gridworld mirroring \thor. See \figref{fig:grid_vision_thor_comparison} for a side-by-side comparison of \thor and our gridworld. This gridworld was implemented primarily in Python with careful caching of data returned from \thor.

\subsubsection{Reward structure.} 
Rewards are provided to each agent individually at every step. These rewards include: (a) $+1$ whenever the lifted object is moved closer, in Euclidean distance, to the goal object than it had been previously in the episode, (b) a constant $-0.01$ step penalty to encourage short trajectories, and (c) a penalty of $-0.02$ whenever the agents action fails. The minimum total reward achievable for a single agent is $-7.5$
corresponding to making only failed actions, while the maximum total reward equals $0.99 \cdot d$ where $d$ is the total number of steps it would take to move the lifted furniture directly to the goal avoiding all obstructions. Our models are trained to maximize the expected discounted cumulative gain with discounting factor $\gamma=0.99$. 

\subsubsection{Optimization and learning hyperparameters.}
For all tasks, we train our agents using reinforcement learning, particularly the popular A3C algorithm~\cite{MnihEtAlPMLR2016}. For {\oldtask}, we follow~\cite{jain2019CVPRTBONE} and additionally use a warm start via imitation learning (DAgger~\cite{RossAISTATS2011}).  When we deploy the coordination loss ({\loss}), we modify the A3C algorithm by replacing the entropy loss with the coordination loss {\loss} defined in~\equref{eq:loss}.
In our experiments we anneal the $\beta$ parameter from a starting value of $\beta = 1$ to a final value of $\beta=0.01$ over the first $5000$ episodes of training. We use an ADAM optimizer with a learning rate of $10^{-4}$, momentum parameters of $0.9$ and $0.999$, with optimizer statistics shared across processes. Gradient updates are performed in an unsynchronized fashion using a HogWild!~style approach \cite{RechtNIPS2011}. Each episode has a maximum length of $250$ total steps per agent. Task-wise details follow:\\
\begin{itemize}
    \item {\task}: Visual agents for {\task} are trained for $500,000$ episodes, across $8$ TITAN V or TITAN X GPUs with $45$ workers and take approximately three days to train.
    
    \item {\gridtask}: Agents for {\gridtask} are trained for 1,000,000 episodes using $45$ workers. Apart from parsing and caching the scene once, gridworld agents do not need to render images. Hence, we  train the agents with only $1$ G4 GPU, particularly the \texttt{g4dn.16xlarge} virtual machine on AWS. Agents (\ie, two) for {\gridtask} take approximately 1 day to train.
    
    \item {\gridtask}-3Agents: Same implementation as above, except that agents (\ie, three) for {\gridtask}-3Agents take approximately 3 days to train. This is due to an increase in the number of forward and backward passes and a CPU bottleneck. Due to the action space blowing up to $|\mathcal{A}|\times|\mathcal{A}|\times|\mathcal{A}| = 2197$ (\vs 169 for two agents), positive rewards become increasingly sparse. This leads to grave inefficiency in training, with no learning for $\sim$500k episodes. To overcome this, we double the positive rewards for the RL formulation for all methods within the three agent setup.
    
    \item {\oldtask}: We adhere to the exact training procedure laid out by Jain~\etal~\cite{jain2019CVPRTBONE}. Visual agents for {\oldtask} are trained for 100,000 episodes with the first 10,000 being warm started with a DAgger-styled imitation learning. Reinforcement learning (A3C) takes over after the warm-start period.
\end{itemize}

\subsubsection{Integration with other MARL methods.} As mentioned in~\secref{sec:related_work}, our contributions are orthogonal to the RL method deployed. Here we give some pointers for integration with a deep Q-Learning and a policy gradient method.\\
\noindent\textbf{QMIX.} While we focus on policy-gradients and QMIX~\cite{rashid2018qmix} uses Q-learning, we can formulate a SYNC for Q-Learning (and QMIX). Analogous to an actor with multiple policies, consider a value head where each agent's Q-function $Q_i$ is replaced by a collection of Q-functions $Q^a_i$ for $a\in A$. Action sampling is done stage-wise, i.e. agents jointly pick a strategy as $\arg\max_a Q_{SYNC}($communications$, a)$, and then individually choose action $\arg\max_{u^i} Q^a_i(\tau^i, u^i)$. These $Q^a_i$ in turn can incorporated into the QMIX mixing network.\\
\noindent\textbf{COMA/MADDPG.} Both these policy gradient algorithms utilize a centralized critic. Since our contributions focus on the actor head, we can directly replace their per-agent policy with our SYNC policies and thus benefit directly from the counterfactual baseline in COMA~\cite{FoersterAAAI2018} or the centralized critic in MADDPG~\cite{LoweNIPS2017}.

\subsection{Quantitative evaluation details} 
\label{sec:quant-eval-extra-details}
\subsubsection{Confidence intervals for metrics reported.} In the main paper, we mentioned that we mark the best performing decentralized method in \textbf{bold} and \hl{highlight it in green} if it has non-overlapping $95$\% confidence intervals. In this {\appendixorsupp}, particularly in \tabref{tab:quant_95conf}, \tabref{tab:quant-furnlift-95conf}, \tabref{tab:mixture_95conf}, and \tabref{tab:cl_study_95conf} we include the 95\% confidence intervals for the metrics reported in \tabref{tab:quant}, \tabref{tab:quant-furnlift}, \tabref{tab:mixture}, and \tabref{tab:cl_study}. 

\subsubsection{Hypotheses on 3-agent \textit{central} method performance.} In \figref{tab:quant} and \secref{sec:quantitative} of the main paper, we mention that the  \textit{central} method performs worse than {\itmethod} for the {\gridtask}-3Agent task. We hypothesize that this is because the \textit{central} method for the -3Agent setup is significantly slower as its actor head has dramatically more parameters requiring more time to train. In numbers -- the \textit{central}'s actor head alone has $D\times|\mathcal{A}|^3$ parameters, where $D$ is the dimensionality of the final representation fed into the actor (please see \figref{fig:central_model} for \textit{central}'s architecture). Note, $D=512$ for our architecture means the \textit{central}'s actor head has $512\cdot 13^3=$1,124,864 parameters. Contrast this to {\itmethod}'s $D\times|\mathcal{A}|\times K$ parameters for a $K$ mixture component. Even for the highest $K$ in the mixture component study (\tabref{tab:mixture}), \ie, $K=13$, this value is $86,528$ parameters. Such a large number of parameters makes learning with the \textit{central} agent slow even after $1$M episodes (this is already $10\times$ more training episodes than used in \cite{jain2019CVPRTBONE}).

\subsubsection{Why MD-SPL instead of SPL?} SPL was introduced in~\cite{anderson2018evaluation} for evaluating single-agent navigational agents, and is defined as follows:
\begin{align}
    \text{SPL} = \frac{1}{N_\text{ep}}\sum_{i=1}^{N_\text{ep}}S_i\frac{l_i}{\max(x_i, l_i)},    
\end{align}
where $i$ denotes an index over episodes, $N_\text{ep}$ equals the number of test episodes, and $S_i$ is a binary indicator for success of episode $i$. Also $x_i$ is the length of the agent's path and $l_i$ is the shortest-path distance from agent's start location to the goal. Directly adopting SPL  isn't pragmatic for two reasons:
\begin{enumerate}[(a)]\compresslist
    \item Coordinating actions \textit{at every timestep} is critical to this multi-agent task. Therefore, the number of actions taken by agents instead of distance (say in meters) should be incorporated in the metric.
    \item Shortest-path distance has been calculated for two agent systems for {\oldtask} \cite{jain2019CVPRTBONE} by finding the shortest path for each agent in a state graph. This can be done effectively for fairly independent agents.
    While each position of the agent corresponds to 4 states (if 4 rotations are possible), each position of the furniture object corresponds to 
    \begin{align}
        \text{\#~\text{States}} =~ & (\# \text{pos. for}~A^1~\text{near obj})\times(\#\text{pos. for}~A^2~\text{near obj})
        \label{eq:states}\\
        & \times (\#\text{rot. for obj}) \times (\#\text{rot. for}~A^1) \times (\#\text{rot. for}~A^2), \nonumber
    \end{align}
    This leads to 404,480 states for an agent-object-agent assembly. We found the shortest path algorithm to be intractable in a state graph of this magnitude. Hence we resort to the closest approximation of Manhattan distance from the object's start position to the goal's position. This is the shortest path, if there were no obstacles for navigation.
\end{enumerate}
Minimal edits to resolve the above two problems lead us to using actions instead of distance, and leveraging Manhattan distance instead of shortest-path distance. This leads us to defining, as described in Section~\secref{sec:metrics} of the main paper, the Manhattan distance based SPL (\text{MDSPL}) as the quantity
\begin{align}
    \text{MDSPL} = \frac{1}{N_\text{ep}}\sum_{i=1}^{N_\text{ep}}S_i\frac{m_i/d_\text{grid}}{\max(p_i, m_i/d_\text{grid})}.    
\end{align}

\subsubsection{Defining additional metrics used for {\oldtask}.} Jain~\etal~\cite{jain2019CVPRTBONE} use two metrics which they refer to as \textit{failed pickups} (picked up, but not `pickupable') and \textit{missed pickups} (`pickupable' but not picked up). `Pickupable' means when the object and agent configurations were valid for a \textsc{PickUp} action.

\subsubsection{Plots for additional metrics.}

\begin{figure}[t]
    \centering
    \includegraphics[width=\textwidth]{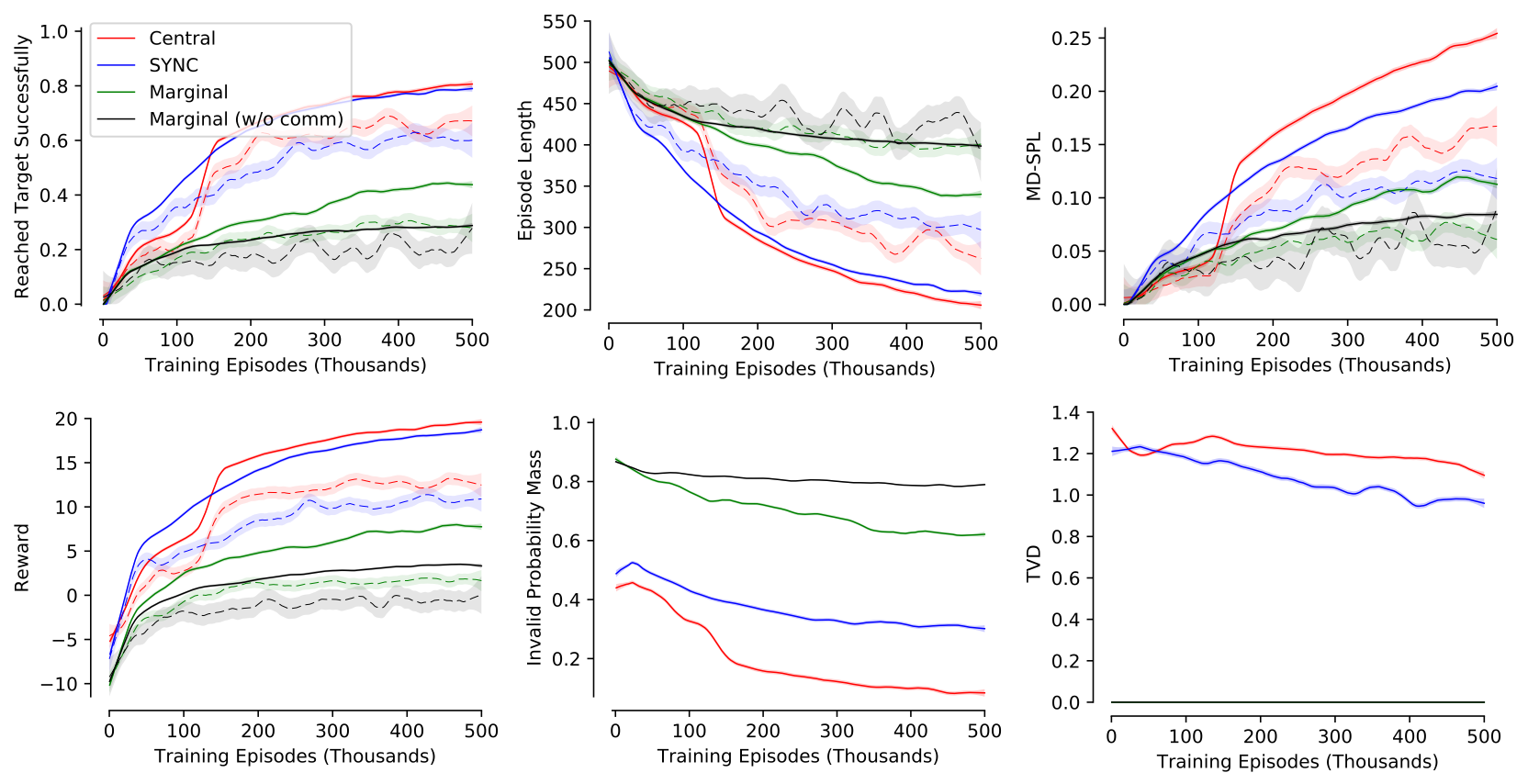}
    \caption{\textbf{Metrics recorded during training for the \task task for various models.} These plots add to the graph shown in \figref{fig:tb-graphs-furnmove}. Here solid lines indicate performance on the training set and dashed lines the performance on the validation set. For the \invalidprob and \lrankdist metrics, only training set values are shown. For the \lrankdist metric the black line (corresponding to the \textit{Marginal (w/o comm)} model completely covers the green line corresponding to the \textit{Marginal} model.}
    \label{fig:all-train-plots-vision-furnmove}
\end{figure}

\begin{figure}[t]
    \centering
    \includegraphics[width=\textwidth]{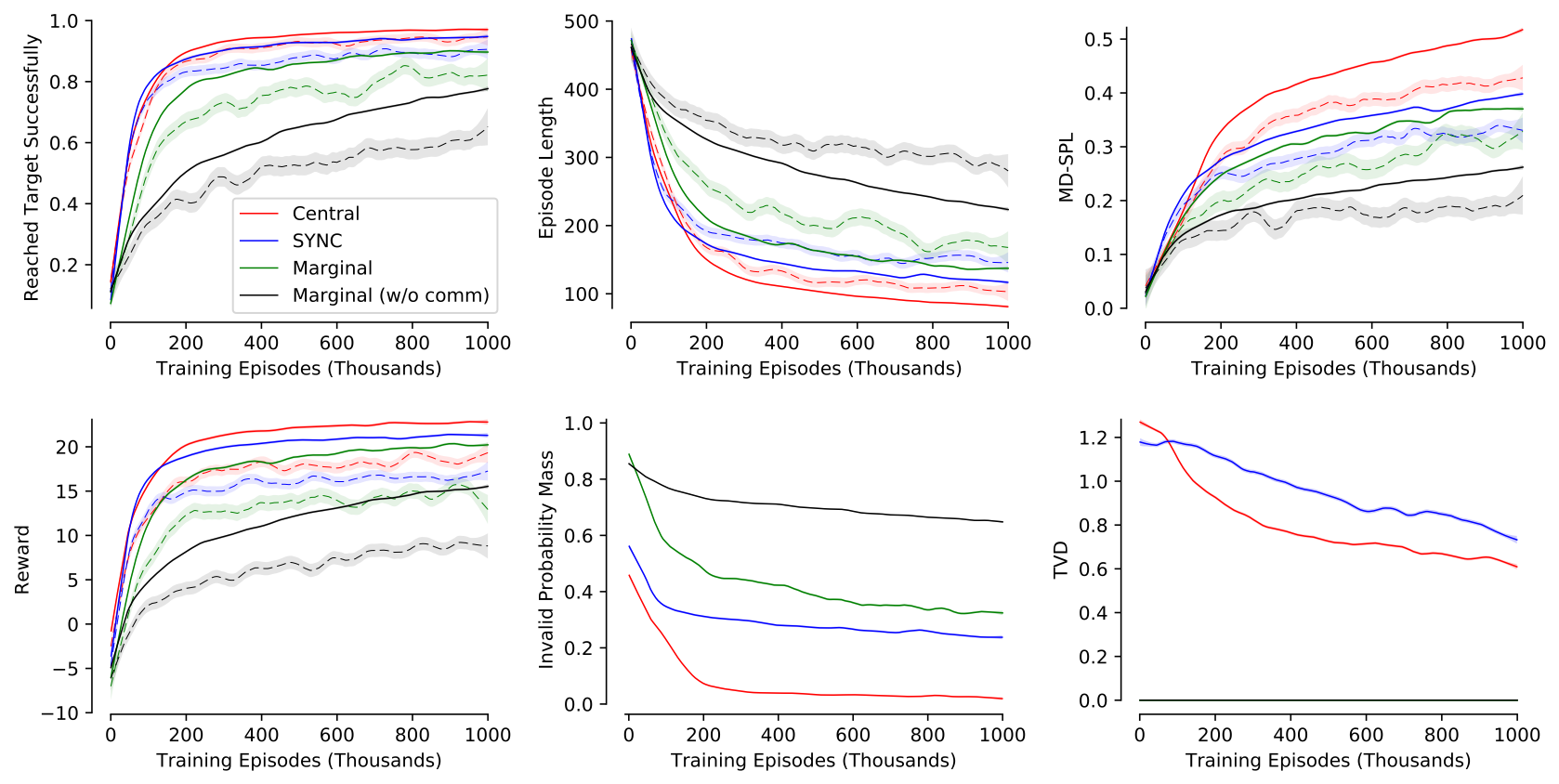}
    \caption{\textbf{Metrics recorded during training for the \gridtask task for various models.} These plots add to the graph shown in \figref{fig:tb-graphs-grid-furnmove}. Here solid lines indicate performance on the training set and dashed lines the performance on the validation set. For the \invalidprob and \lrankdist metrics, only training set values are shown. For the \lrankdist metric the black line (corresponding to the \textit{Marginal (w/o comm)} model completely covers the green line corresponding to the \textit{Marginal} model.}
    \label{fig:all-train-plots-grid-furnmove}
\end{figure}

\begin{figure}[t]
    \centering
    \includegraphics[width=\textwidth]{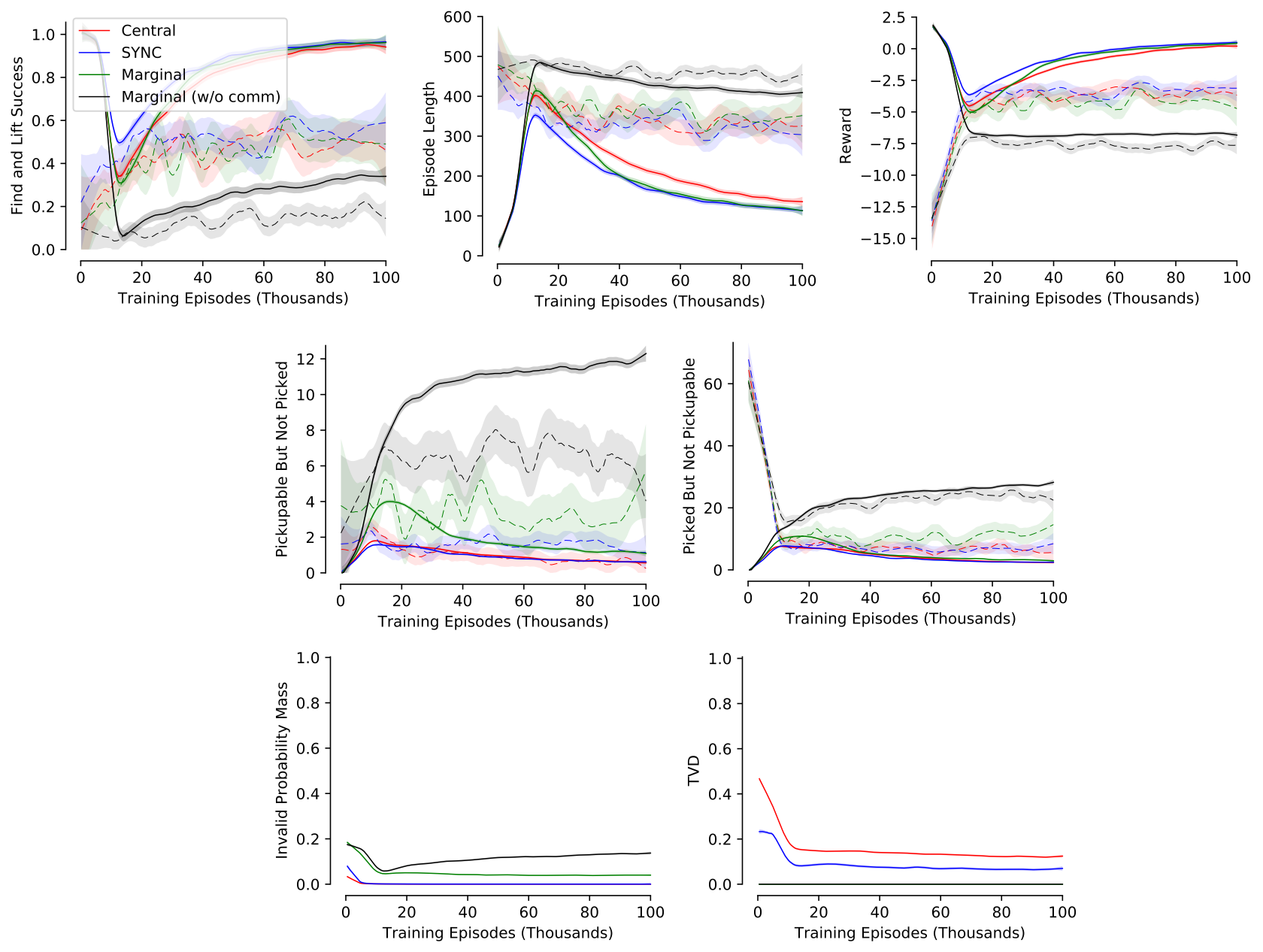}
    \caption{\textbf{Metrics recorded during training for the \oldtask task for various models.} These plots add to the graph shown in \figref{fig:tb-graphs-furnlift}. Notice that we have included plots corresponding to the \textit{failed pickups} (picked up, but not `pickupable') and \textit{missed pickups} (`pickupable' but not picked up) metrics described in \secref{sec:quant-eval-extra-details}. Solid lines indicate performance on the training set and dashed lines the performance on the validation set. For the \invalidprob and \lrankdist metrics, only training set values are shown. For the \lrankdist metric the black line (corresponding to the \textit{Marginal (w/o comm)} model completely covers the green line corresponding to the \textit{Marginal} model.}
    \label{fig:all-train-plots-vision-furnlift}
\end{figure}

See Fig. \ref{fig:all-train-plots-vision-furnmove}, \ref{fig:all-train-plots-grid-furnmove}, and \ref{fig:all-train-plots-vision-furnlift} for plots of additional metric recorded during training for the \task, \gridtask, and \oldtask tasks. \figref{fig:all-train-plots-vision-furnlift} in particular shows how the \textit{failed pickups} and \textit{missed pickups} metrics described above are substantially improved when using our \method models.

\subsubsection{Additional 3-agent experiments.}

In the main paper we present results when training \itmethod, \textit{marginal}, and \textit{central} models to complete the 3-agent \gridtask task. We have also trained the same methods to complete the (visual) 3-agent \task task. Rendering and simulating 3-agent interactions in \thor is computationally taxing. For this reason we  trained our \itmethod and \textit{central} models for 300k episodes instead of the 500k episodes we used when  training 2-agent models. As it showed no training progress, we also stopped the \textit{marginal} model's training after 100k episodes. Training until 300k episodes took approximately four days using eight 12GB GPUs ($\sim 768$ GPU hours per model).

After training, the \itmethod, \textit{marginal}, and \textit{central} obtained a test-set success rate of $23.2 \pm 2.6$\%, $0.0 \pm 0.0$\%, and $12.4 \pm 2.0$\% respectively. These results mirror those of the 3-agent \gridtask task from the main paper. Particularly, both the \itmethod and \textit{central} models train to reasonable success rates but the \textit{central} model  actually performs worse than the \itmethod model. A discussion of our hypothesis for why this is the case can be found earlier in this section. In terms of our other illustrative metrics, our \itmethod, \textit{marginal}, and \textit{central} respectively obtain MDSPL values of $0.029$, $0.0$, and $0.012$, and \invalidprob values of $0.336$, $0.854$, and $0.132$.

\begin{table}[t]
\caption{95\% confidence intervals included in addition to~\tabref{tab:quant}, evaluating methods on {\task}, {\gridtask}, and {\gridtask-3Agents}. For legend details, see \tabref{tab:quant}.
} \label{tab:quant_95conf}
\setlength{\tabcolsep}{6pt}
\centering
\resizebox{\linewidth}{!}{%
\begin{tabular}{ccccccc}
\hline
\rowcolor[HTML]{EFEFEF} 
\textbf{Methods} &
\textbf{MD-SPL} $\uparrow$ &
\textbf{Success} $\uparrow$&
\textbf{Ep len} $\downarrow$&
\textbf{\begin{tabular}[c]{@{}c@{}}Final \\ dist\end{tabular}} $\downarrow$& \textbf{\begin{tabular}[c]{@{}c@{}}Invalid \\ prob.\end{tabular}} $\downarrow$&
\textbf{TVD} $\updownarrow$\\ 
\hline
\multicolumn{7}{c}{{\task} (ours)} \\ \hline
\multirow{2}{*}{Marginal w/o comm~\cite{jain2019CVPRTBONE}} & 0.032 & 0.164 & 224.1 & 2.143 & 0.815 & 0 \\
& ($\pm$0.007) & ($\pm$0.023) & ($\pm$2.031) & ($\pm$0.104) & ($\pm$0.005) & ($\pm$0)\\
\multirow{2}{*}{Marginal~\cite{jain2019CVPRTBONE}} & 0.064 & 0.328 & 194.6 & 1.828 & 0.647 & 0 \\
& ($\pm$0.008) & ($\pm$0.029) & ($\pm$2.693) & ($\pm$0.105) & ($\pm$0.010) & ($\pm$0)\\
\multirow{2}{*}{{\method}} & \hl{\textbf{0.114}} & \hl{\textbf{0.587}} & \hl{\textbf{153.5}} & \hl{\textbf{1.153}} & \hl{\textbf{0.31}} & 0.474 \\
& ($\pm$0.009) & ($\pm$0.031) & ($\pm$2.870) & ($\pm$0.089) & ($\pm$0.004) & ($\pm$0.005)\\
\hdashline
\multirow{2}{*}{Central$^\dagger$} & 0.161 & 0.648 & 139.8 & 0.903 & 0.075 & 0.543 \\
& ($\pm$0.012) & ($\pm$0.030) & ($\pm$2.958) & ($\pm$0.076) & ($\pm$0.006) & ($\pm$0.006)\\
\hline

\multicolumn{7}{c}{{\gridtask} (ours)} \\ \hline
\multirow{2}{*}{Marginal w/o comm~\cite{jain2019CVPRTBONE}} & 0.111 & 0.484 & 172.6 & 1.525 & 0.73 & 0 \\
& ($\pm$0.012) & ($\pm$0.031) & ($\pm$2.825) & ($\pm$0.121) & ($\pm$0.008) & ($\pm$0) \\
\multirow{2}{*}{Marginal~\cite{jain2019CVPRTBONE}} & 0.218 & 0.694 & 120.1 & 0.960 & 0.399 & 0 \\
& ($\pm$0.015) & ($\pm$0.029) & ($\pm$2.974) & ($\pm$0.100) & ($\pm$0.011) & ($\pm$0) \\
\multirow{2}{*}{{\method}} & \textbf{0.228} & \hl{\textbf{0.762}} & \textbf{110.4} & \hl{\textbf{0.711}} & \hl{\textbf{0.275}} & 0.429 \\
& ($\pm$0.014) & ($\pm$0.026) & ($\pm$2.832) & ($\pm$0.076) & ($\pm$0.005) & ($\pm$0.005) \\
\hdashline
\multirow{2}{*}{Central$^\dagger$} & 0.323 & 0.818 & 87.7 & 0.611 & 0.039 & 0.347 \\
& ($\pm$0.016) & ($\pm$0.024) & ($\pm$2.729) & ($\pm$0.067) & ($\pm$0.004) & ($\pm$0.006) \\
\hline

\multicolumn{7}{c}{{\gridtask}-3Agents (ours)} \\ \hline
\multirow{2}{*}{Marginal~\cite{jain2019CVPRTBONE}} & 0 & 0 & 250.0 & 3.564 & 0.823 & 0\\
& ($\pm$0) & ($\pm$0) & ($\pm$0) & ($\pm$0.111) & ($\pm$0) & ($\pm$0) \\
\multirow{2}{*}{{\method}} & \hl{\textbf{0.152}} & \hl{\textbf{0.578}} & \hl{\textbf{149.1}} & \hl{\textbf{1.05}} & \hl{\textbf{0.181}} & 0.514\\
& ($\pm$0.012) & ($\pm$0.031) & ($\pm$6.020) & ($\pm$0.091) & ($\pm$0.006) & ($\pm$0.009) \\
\hdashline
\multirow{2}{*}{Central$^\dagger$} & 0.066 & 0.352 & 195.4 & 1.522 & 0.138 & 0.521\\
& ($\pm$0.008) & ($\pm$0.03) & ($\pm$5.200) & ($\pm$0.099) & ($\pm$0.005) & ($\pm$0.006) \\
\hline

\end{tabular}
}
\end{table}

\begin{table}[t]
\caption{
95\% confidence intervals included in addition to~\tabref{tab:quant-furnlift}, evaluating methods on {\oldtask}. \textit{Marginal} and {\itmethod} perform equally well, and mostly lie within confidence intervals of each other. \textit{Invalid prob.} and \textit{failed pickups} metrics for {\itmethod} have non-overlapping confidence bounds (\hl{lighted in green}). For more details on the legend, see \tabref{tab:quant}.
} 
\label{tab:quant-furnlift-95conf}
\setlength{\tabcolsep}{3pt}
\centering
\resizebox{\linewidth}{!}{
\begin{tabular}{ccccccccc}
\hline
\rowcolor[HTML]{EFEFEF} 
\textbf{Methods} &
\textbf{MD-SPL} $\uparrow$ &
\textbf{Success} $\uparrow$&
\textbf{Ep len} $\downarrow$&
\textbf{\begin{tabular}[c]{@{}c@{}}Final \\ dist\end{tabular}} $\downarrow$& 
\textbf{\begin{tabular}[c]{@{}c@{}}Invalid \\ prob.\end{tabular}} $\downarrow$&
\textbf{TVD} $\updownarrow$&
\textbf{\begin{tabular}[c]{@{}c@{}}Failed \\pickups\end{tabular}} $\downarrow$& 
\textbf{\begin{tabular}[c]{@{}c@{}}Missed \\pickups\end{tabular}} $\downarrow$\\
\hline
\multicolumn{9}{c}{{\oldtask}~\cite{jain2019CVPRTBONE} (`constrained' setting with no implicit communication)} \\ \hline
\multirow{2}{*}{\begin{tabular}[c]{@{}c@{}}Marginal w/o comm~\cite{jain2019CVPRTBONE}\end{tabular}} & 0.029 & 0.15 & 229.5 & 2.455 & 0.11 & 0 & 25.219 & 6.501 \\
& ($\pm$0.007) & ($\pm$0.022) & ($\pm$3.482) & ($\pm$0.105) & ($\pm$0.004) & ($\pm$0) & ($\pm$1.001) & ($\pm$0.784) \\
\multirow{2}{*}{Marginal~\cite{jain2019CVPRTBONE}} & \textbf{0.145} & \textbf{0.449} & \textbf{174.1} & 2.259 & 0.042 & 0 & 8.933 & 1.426 \\
& ($\pm$0.016) & ($\pm$0.031) & ($\pm$5.934) & ($\pm$0.094) & ($\pm$0.003) & ($\pm$0) & ($\pm$0.867) & ($\pm$0.284) \\
\multirow{2}{*}{{\method}} & 0.139 & 0.423 & 176.9 & \textbf{2.228} & \hl{\textbf{0}} & 0.027 & \hl{\textbf{4.873}} & \textbf{1.048} \\
& ($\pm$0.016) & ($\pm$0.031) & ($\pm$5.939) & ($\pm$0.083) & ($\pm$0) & ($\pm$0.002) & ($\pm$0.453) & ($\pm$0.192) \\
\hdashline
\multirow{2}{*}{Central$^\dagger$} & 0.145 & 0.453 & 172.3 & 2.331 & 0 & 0.059 & 5.145 & 0.639 \\
& ($\pm$0.016) & ($\pm$0.031) & ($\pm$5.954) & ($\pm$0.088) & ($\pm$0) & ($\pm$0.002) & ($\pm$0.5) & ($\pm$0.164) \\
\hline
\end{tabular}
}
\end{table}

\begin{table}[t]
\caption{95\% confidence intervals included in addition to \tabref{tab:mixture} by varying number of components in SYNC-policies for \task.
}
\label{tab:mixture_95conf}
\centering
\setlength{\tabcolsep}{6pt}
\resizebox{0.85\linewidth}{!}{%
\begin{tabular}{c@{\hskip6pt}c@{\hskip6pt}c@{\hskip6pt}c@{\hskip6pt}c@{\hskip6pt}c@{\hskip6pt}c} 
\hline
\rowcolor[HTML]{EFEFEF} 
\textbf{$K$ in {\method}}  &
\textbf{MD-SPL} $\uparrow$ &
\textbf{Success} $\uparrow$&
\textbf{Ep len} $\downarrow$&
\textbf{\begin{tabular}[c]{@{}c@{}}Final \\ dist\end{tabular}} $\downarrow$& \textbf{\begin{tabular}[c]{@{}c@{}}Invalid \\ prob.\end{tabular}} $\downarrow$&
\textbf{TVD} $\updownarrow$\\ \hline
\multicolumn{7}{c}{\task} \\ \hline
\multirow{2}{*}{1 component} & 0.064 & 0.328 & 194.6 & 1.828 & 0.647 & 0 \\
& ($\pm$0.004) & ($\pm$0.019) & ($\pm$2.833) & ($\pm$0.105) & ($\pm$0.002) & ($\pm$0) \\
\multirow{2}{*}{2 components} & 0.084 & 0.502 & 175.5 & 1.227 & \textbf{0.308} & 0.206 \\
& ($\pm$0.008) & ($\pm$0.031) & ($\pm$5.321) & ($\pm$0.091) & ($\pm$0.004) & ($\pm$0.004) \\
\multirow{2}{*}{4 components} & 0.114 & 0.569 & 154.1 & \textbf{1.078} & 0.339 & 0.421 \\
& ($\pm$0.009) & ($\pm$0.031) & ($\pm$5.783) & ($\pm$0.083) & ($\pm$0.004) & ($\pm$0.005) \\
\multirow{2}{*}{13 components} & \textbf{0.114} & \textbf{0.587} & \textbf{153.5} & 1.153 & 0.31 & 0.474 \\
& ($\pm$0.009) & ($\pm$0.031) & ($\pm$5.739) & ($\pm$0.089) & ($\pm$0.004) & ($\pm$0.005) \\
\hline
\end{tabular}%
}
\end{table}

\begin{table}[t]
\caption{
95\% confidence intervals included in addition to \tabref{tab:cl_study}, ablating coordination loss on \textit{marginal}~\cite{jain2019CVPRTBONE}, \itmethod, and \textit{central} methods. $^\dagger$denotes that a centralized agent serve only as an upper bound to decentralized methods.
}
\label{tab:cl_study_95conf}
\centering
\setlength{\tabcolsep}{4pt}
\resizebox{0.95\linewidth}{!}{%
\begin{tabular}{cccccccc} 
\hline\rowcolor[HTML]{EFEFEF} 
\textbf{Method}  &
\textbf{{\loss}} &
\textbf{MD-SPL} $\uparrow$ &
\textbf{Success} $\uparrow$&
\textbf{Ep len} $\downarrow$&
\textbf{\begin{tabular}[c]{@{}c@{}}Final \\ dist\end{tabular}} $\downarrow$& \textbf{\begin{tabular}[c]{@{}c@{}}Invalid \\ prob.\end{tabular}} $\downarrow$&
\textbf{TVD} $\updownarrow$\\ \hline
\multicolumn{8}{c}{ \textit{\task} } \\ 
\hline
\multirow{2}{*}{Marginal} & \multirow{2}{*}{{\color{red} \ding{55}}} &  0.064 & 0.328 & 194.6 & 1.828 & 0.647 & 0\\
& &  ($\pm$0.008) & ($\pm$0.029) & ($\pm$5.385) & ($\pm$0.105) & ($\pm$0.01) & ($\pm$0.0) \\
\multirow{2}{*}{Marginal} & \multirow{2}{*}{{\color{green} \ding{51}}} & 0.015 & 0.099 & 236.9 & 2.134 & 0.492 & 0\\
& &  ($\pm$0.004) & ($\pm$0.019) & ($\pm$2.833) & ($\pm$0.105) & ($\pm$0.002) & ($\pm$0.0) \\
\multirow{2}{*}{\method} & \multirow{2}{*}{{\color{red} \ding{55}}} &  0.091 & 0.488 & 170.3 & 1.458 & 0.47 & 0.36\\
& &  ($\pm$0.008) & ($\pm$0.031) & ($\pm$5.665) & ($\pm$0.104) & ($\pm$0.008) & ($\pm$0.008) \\
\multirow{2}{*}{\method} & \multirow{2}{*}{{\color{green} \ding{51}}} &  0.114 & 0.587 & 153.5 & 1.153 & 0.31 & 0.474\\
& &  ($\pm$0.009) & ($\pm$0.031) & ($\pm$5.739) & ($\pm$0.089) & ($\pm$0.004) & ($\pm$0.005) \\
\hdashline
\multirow{2}{*}{Central$^\dagger$} & \multirow{2}{*}{{\color{red} \ding{55}}} &  0.14 & 0.609 & 146.9 & 1.018 & 0.155 & 0.6245\\
& &  ($\pm$0.011) & ($\pm$0.03) & ($\pm$5.895) & ($\pm$0.084) & ($\pm$0.006) & ($\pm$0.005) \\
\multirow{2}{*}{Central$^\dagger$} & \multirow{2}{*}{{\color{green} \ding{51}}} &  0.161 & 0.648 & 139.8 & 0.903 & 0.075 & 0.543\\
& &  ($\pm$0.012) & ($\pm$0.03) & ($\pm$5.915) & ($\pm$0.076) & ($\pm$0.006) & ($\pm$0.006) \\
\hline
\end{tabular}
}
\end{table}

\subsection{Qualitative evaluation details and a statistical analysis of learned communication}
\label{sec:qualitative-extra-details}

\subsubsection{Discussion of our qualitative video.}
\begin{figure*}[t]
    \centering
    \includegraphics[width=\linewidth]{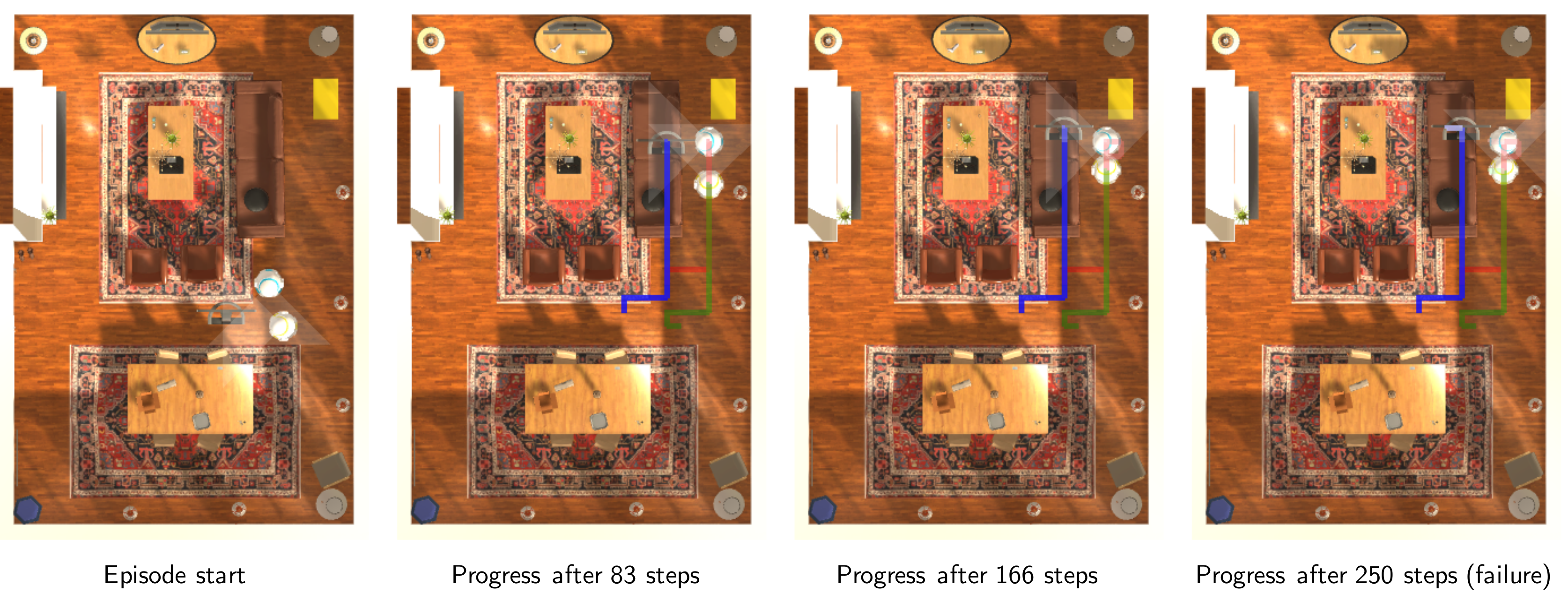}
    \caption{Clip A trajectory summary. The marginal agents quickly get stuck in a narrow area between a sofa and the wall and fail to make progress.}
    \label{fig:clipa}
\end{figure*}

\begin{figure*}[t]
    \centering
    \includegraphics[width=\linewidth]{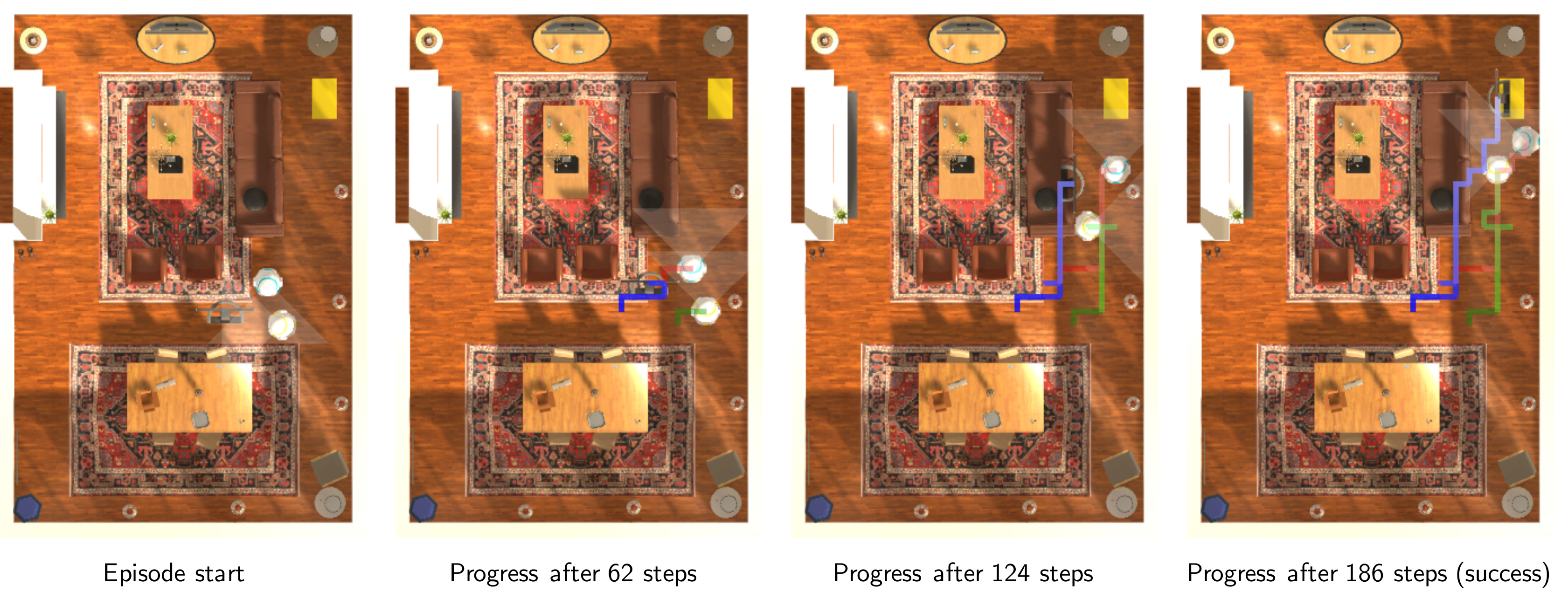}
    \caption{Clip B trajectory summary. The SYNC agents successfully navigate the TV to the goal location without getting stuck in the narrow corridor.}
    \label{fig:clipb}
\end{figure*}

We include a video of policy roll-outs in the supplementary material. This includes  four clips, each corresponding to the rollout on a test scene of one of our models trained to complete the \task task.\\
\noindent\textbf{Clip A.} \textit{Marginal} agents attempt to move the TV to the goal but get stuck in a narrow corridor as they struggle to successfully coordinate their actions. The episode is considered a failure as the agents do not reach the goal in the allotted 250 timesteps. A top-down summary of this trajectory is included in~\figref{fig:clipa}.\\
\noindent\textbf{Clip B.} Unlike the \textit{marginal} agents from Clip A., in this clip two \textit{SYNC} agents successfully coordinate actions and move the TV to the goal location in $186$ steps. A top-down summary of this trajectory is included in~\figref{fig:clipb}.\\
\noindent\textbf{Clip C.} Here we show \textit{SYNC} agents completing the \gridtask in a test scene (the same scene and initial starting positions as in Clip A and Clip B). The agents complete the task in 148 timesteps even after an initial search in the incorrect direction.\\
\noindent\textbf{Clip D (contains audio).} This clip is an attempt to \textit{experience} what agents `hear.' The video for this clip is the same as Clip B showing the \itmethod method. The audio is a rendering of the communication between agents in the reply stage. Particularly, we discretize the $[0,1]$ value associated with the first reply weight of each agent into 128 evenly spaced bins corresponding to the 128 notes on a MIDI keyboard (0 corresponding to a frequency of $\sim$8.18 Hz and 127 to $\sim$12500 Hz). Next, we post-process the audio so that the communication from the agents is played on different channels (stereo) and has the Tech Bass tonal quality. As a result, the reader can experience what agent 1 hears (\ie, agent 2's reply weight) via the left earphone/speaker and what agent 2 hears (\ie, agent 1's reply weight) via the right speaker. In addition to the study in \secref{sec:qualitative} and \secref{sec:supp-comm}, we notice a  higher pitch/frequency for the agent which is passing. We also notice lower pitches for \textsc{MoveWithObject} and \textsc{MoveObject} actions. \\

\subsubsection{Joint policy summaries.} 
\begin{figure}[t]
    \centering
    \includegraphics[width=0.9\linewidth]{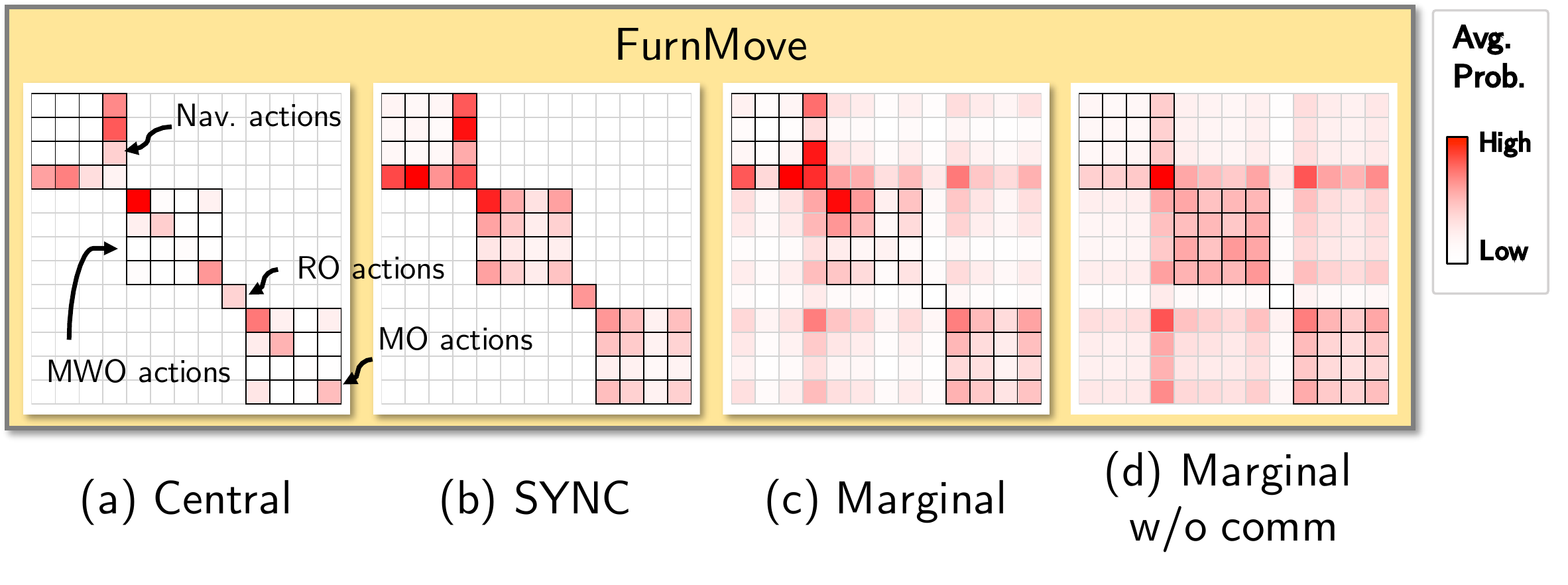}
    \includegraphics[width=0.9\linewidth]{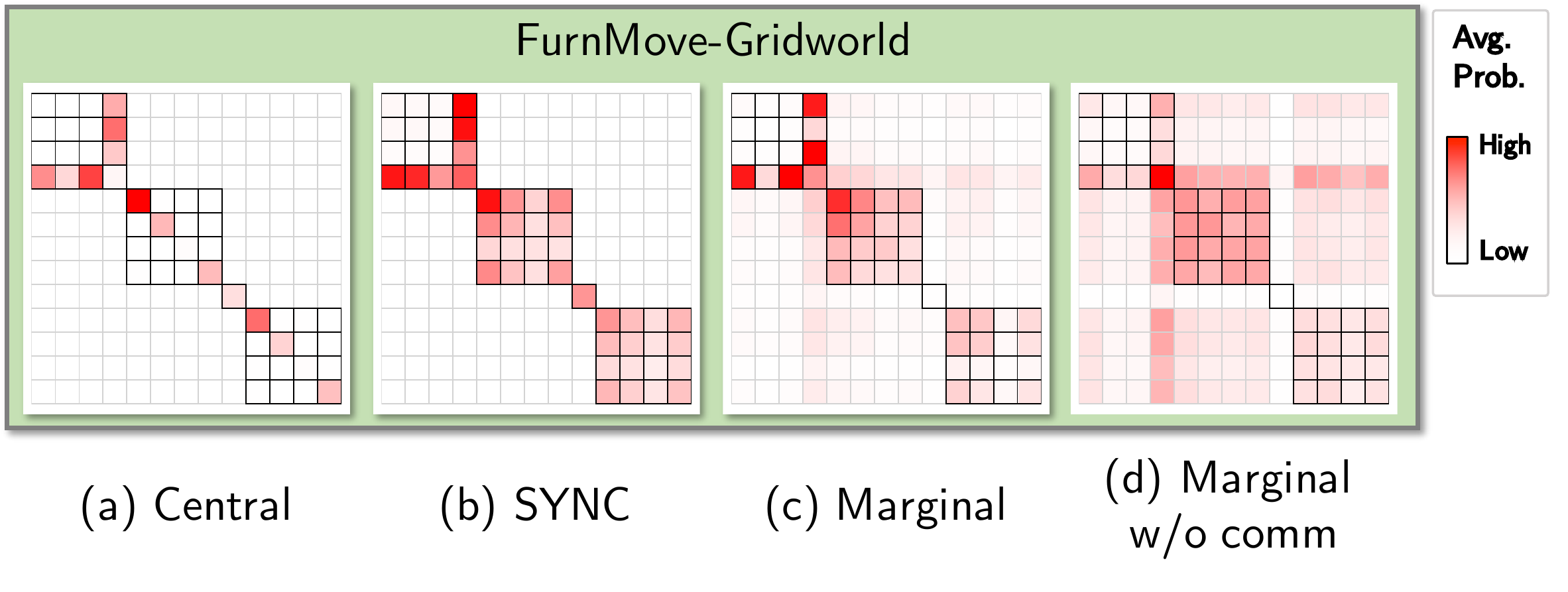}
    \caption{\textbf{Additional results for~\figref{fig:matrices}.} Joint policy summaries for all methods for both {\task} and {\gridtask}.}
    \label{fig:matrices-more}
\end{figure}

These provide a way to visualize the effective joint distribution that each method captures. For each episode in the test set, we log each multi-action attempted by a method. We average over steps in the episode to obtain a matrix (which sums to one). Afterwards, we average these matrices (one for each episode) to create a \textit{joint policy summary} of the method for the entire test set. This two-staged averaging prevents the snapshot from being skewed towards actions enacted in longer (failed or challenging) episodes. In the main paper, we included snapshots for {\task} in~\figref{fig:matrices}. In \figref{fig:matrices-more} we include additional visualizations for all methods including (\textit{Marginal w/o comm} model) for {\task} and {\gridtask}. %

\subsubsection{Communication analysis.}
\label{sec:supp-comm}
As shown in Fig.~\ref{fig:matrices_communication_analysis} and discussed in Sec.~\ref{sec:qualitative}, there is very strong qualitative evidence suggesting that our agents use their talk and reply communication channels to explicitly relay their intentions and coordinate their actions. We now produce a statistical, quantitative, evaluation of this phenomenon by fitting multiple logistic regression models where we attempt to predict, from the agents communications, certain aspects of their environment as well as their future actions. In particular, we run 1000 episodes on our test set using our mixture model in the visual testbed. This produces a dataset of 159,380 observations where each observation records, for a single step by both agents at time $t$:
\begin{enumerate}[(a)]
    \item The two weights $p^{1}_{\text{talk}, t}, p^{2}_{\text{talk}, t}$ where $p^{i}_{\text{talk}, t}$ is the weight agent $A^i$ assigns to the first symbol in the ``talk'' vocabulary.
    \item The two weights $p^{1}_{\text{reply}, t}, p^{2}_{\text{reply}, t}$ where $p^{i}_{\text{reply}, t}$ is the weight agent $A^i$ assigns to the first symbol in the ``reply'' vocabulary.
    \item The two values $\texttt{tv}^i_t\in\{0,1\}$ where $\texttt{tv}^i_t$ equals 1 if and only if agent $A^i$ sees the TV at timestep $t$ (before taking its action).
    \item The two values $\texttt{WillPass}^i_t\in\{0,1\}$ where $\texttt{WillPass}^i_t$ equals 1 if and only if agent $i$ ends up choosing to take the \textsc{Pass} action at time $t$ (\ie, after finishing communication).
    \item The two values $\texttt{WillMWO}^i_t\in\{0,1\}$ where $\texttt{WillMWO}^i_t$ equals 1 if and only if agent $i$ ends up choosing to take some \textsc{MoveWithObject} action at time $t$.
\end{enumerate}
In the following we will drop the subscript $t$ and consider the above quantities as random samples drawn from the distribution of possible steps taken by our agents in randomly initialized trajectories. As $A^1$ and $A^2$ share almost all of their parameters they are, essentially, interchangeable. Because of this our following analysis will be solely taking the perspective of agent $A^1$, similar results hold for $A^2$. We consider fitting the three models:
\begin{align}
    \sigma^{-1} P(\texttt{tv}^1_t =1) &= \beta_{\text{tv}} + \beta^1_{\text{talk, tv}}\cdot p^{1}_\text{talk} \label{eq:tv_vis}\\
    &\quad + \beta^1_{\text{reply, tv}}\cdot p^1_{\text{talk}}, \nonumber\\
    &\quad + \beta^1_{\text{talk*reply, tv}}\cdot p^1_{\text{talk}}\cdot p^1_{\text{reply}}, \nonumber
\end{align}
\begin{align}
    \sigma^{-1} P(\texttt{WillPass}^1=1) &= \beta_{\text{pass}} \label{eq:willpass}\\
    &\quad + \sum_{i=1}^2\beta^i_{\text{talk, pass}}\cdot p^{i}_\text{talk} \nonumber \\
    &\quad + \sum_{i=1}^2\beta^i_{\text{reply, pass}}\cdot p^{i}_\text{reply}, \text{and} \nonumber
\end{align}
\begin{align}
    \sigma^{-1} P(\texttt{WillMWO}^1=1) &= \beta_{\text{MWO}} \label{eq:willmwo}\\
    &\quad + \sum_{i=1}^2\beta^i_{\text{talk, MWO}}\cdot p^{i}_\text{talk} \nonumber \\
    &\quad + \sum_{i=1}^2\beta^i_{\text{reply, MWO}}\cdot p^{i}_\text{reply}, \nonumber
\end{align}
where $\sigma$ is the usual logistic function. Here \equref{eq:tv_vis} attempts to determine the relationship between what $A^1$ communicates and whether or not $A^1$ is currently seeing the TV, \equref{eq:willpass} probes whether or not any communication symbol is associated with $A^1$ choosing to take a \textsc{Pass} action, and finally \equref{eq:willmwo} considers whether or not $A^1$ will choose to take a \textsc{MoveWithObject} action. We fit each of the above models using the \texttt{glm} function in the R programming language \cite{RCoreTeam2019}. Moreover, we compute confidence intervals for our coefficient values using a robust bootstrap procedure. Fitted parameter values can be found in \tabref{table:glm-fits}.

From \tabref{table:glm-fits} we draw several conclusions. First, in our dataset, there is a somewhat complex association between agent $A^1$ seeing the TV and the communication symbols it sends. In particular, for a fixed reply weight $p^1_{\text{reply}} < 0.821$, a larger value of $p^{1}_\text{talk}$ is associated with higher odds of the TV being visible to $A^1$ but if $p^1_{\text{reply}} > 0.821$ then larger values of $p^{1}_\text{talk}$ are associated with smaller odds of the TV being visible. When considering whether or not $A^1$ will pass, the table shows that this decision is strongly associated with the value of $p^{2}_\text{reply}$ where, given fixed values for the other talk and reply weights, $p^{2}_\text{reply}$ being larger by a unit of 0.1 is associated with  $2.7\times$ larger odds of $A^1$ taking the pass action. This suggests the interpretation of a large value of $p^{2}_\text{reply}$ as $A^2$ communicating that it wishes $A^1$ to pass so that $A^1$ may perform a single-agent navigation action to reposition itself. Finally, when considering the fitted values corresponding to Eq.~\eqref{eq:willmwo} we see that while the talk symbols communicated by the agents are weakly related with whether or not $A^1$ takes a \textsc{MoveWithObject} action, the reply symbols are associated with coefficients with an order of magnitude larger values. In particular, assuming all other communication values are  fixed, a smaller value of either $p^{1}_\text{reply}$ or $p^{2}_\text{reply}$ is associated with  substantially larger odds of $A^1$ choosing a \textsc{MoveWithObject} action. This suggests interpreting an especially small value of $p^{i}_\text{reply}$ as agent $A^i$ indicating its readiness to move the object.

\begin{table}[t]
\caption{Estimates, and corresponding robust bootstrap standard errors, for the parameters of communication analysis (\secref{sec:supp-comm}).}\label{table:glm-fits}
\centering
\begin{footnotesize}
\hspace{-2.5mm}
\begin{tabular}{|c|ccccc|}\hline
  &$\beta_{\text{tv}}$ & $\beta^1_{\text{talk, tv}}$ & $\beta^1_{\text{reply, tv}}$ & $\beta^1_{\text{talk*reply, tv}}$&  - \\ \hline
  Est.
  & -2.62 & 6.93 & 3.35 & -8.44 & - \\
  SE &  0.33 & 0.52 & 0.38 & 0.62 & - \\ \hline\hline
  & $\beta_{\text{pass}}$ & $\beta^1_{\text{talk, pass}}$ & $\beta^2_{\text{talk, pass}}$ & $\beta^1_{\text{reply, pass}}$ & $\beta^2_{\text{reply, pass}}$ \\ \hline
  Est.
  & -7.55 & 2.69 & -2.2 & -1.72 & 9.98 \\
  SE & 0.09 & 0.09 & 0.08 & 0.07 & 0.11 \\ \hline\hline
  &$\beta_{\text{MWO}}$ & $\beta^1_{\text{talk, MWO}}$ & $\beta^2_{\text{talk, MWO}}$ & $\beta^1_{\text{reply, MWO}}$ & $\beta^2_{\text{reply, MWO}}$ \\ \hline
  Est.
  & 2.71 & 0.39 & 0.28 & -3.34 & -3.37 \\
  SE & 0.05 & 0.06 & 0.06 & 0.06 & 0.06 \\ \hline
\end{tabular}
\end{footnotesize}
\end{table}

\end{document}